%% file: main.tex
\documentclass[manuscript,screen]{acmart}
\usepackage{bm}
\usepackage{enumitem}
\usepackage{amsthm}
\usepackage{wrapfig}
\usepackage{pifont}
\usepackage{multirow}
\usepackage{pifont}  
\newcommand{\cmark}{\ding{51}} 
\newcommand{\xmark}{\ding{55}}
\newcommand{\omark}{\ding{119}}
\usepackage{fancyhdr}
\theoremstyle{definition}
\newtheorem{definition}{Definition}
\AtBeginDocument{%
  \providecommand\BibTeX{{%
    \normalfont B\kern-0.5em{\scshape i\kern-0.25em b}\kern-0.8em\TeX}}}

\setcopyright{acmlicensed}
\copyrightyear{2018}
\acmYear{2018}
\acmDOI{XXXXXXX.XXXXXXX}
\acmConference[Conference acronym 'XX]{Make sure to enter the correct
  conference title from your rights confirmation emai}{June 03--05,
  2018}{Woodstock, NY}
\acmISBN{978-1-4503-XXXX-X/18/06}

\begin{document}
\title{A Systematic Survey of Electronic Health Record Modeling: From Deep Learning Approaches to Large Language Models}

\author{Weijieying Ren}
\email{wjr5337@psu.edu}
\affiliation{%
  \institution{Information Sciences and Technology, The Pennsylvania State University}
  \city{State College}
  \state{Pennsylvania}
  \country{USA}
  \postcode{16803}
}

\author{Jingxi Zhu}
\email{wjr5337@psu.edu}
\affiliation{%
  \institution{Information Sciences and Technology, The Pennsylvania State University}
  \city{State College}
  \state{Pennsylvania}
  \country{USA}
  \postcode{16803}
}
\author{Zehao Liu}
\email{wjr5337@psu.edu}
\affiliation{%
  \institution{Information Sciences and Technology, The Pennsylvania State University}
  \city{State College}
  \state{Pennsylvania}
  \country{USA}
  \postcode{16803}
}

\author{Tianxiang Zhao}
\affiliation{%
  \institution{Information Sciences and Technology, The Pennsylvania State University}
  \city{State College}
  \country{USA}}
\email{tkz5084@psu.edu}

\author{Vasant Honavar}
\affiliation{%
  \institution{Information Sciences and Technology, The Pennsylvania State University}
  \city{State College}
  \country{USA}
}

\renewcommand{\shortauthors}{Trovato and Tobin, et al.}

\begin{abstract}
Artificial intelligence has demonstrated significant potential in transforming healthcare through the analysis and modeling of electronic health records (EHRs). However, the inherent heterogeneity, temporal irregularity, and domain-specific nature of EHR data present unique challenges that differ fundamentally from those in vision and natural language tasks. This survey offers a comprehensive overview of recent advancements at the intersection of deep learning, large language models (LLMs), and EHR modeling. We introduce a unified taxonomy that spans five key design dimensions: data-centric approaches, neural architecture design, learning-focused strategies, multimodal learning, and LLM-based modeling systems. Within each dimension, we review representative methods addressing data quality enhancement, structural and temporal representation, self-supervised learning, and integration with clinical knowledge. We further highlight emerging trends such as foundation models, LLM-driven clinical agents, and EHR-to-text translation for downstream reasoning. Finally, we discuss open challenges in benchmarking, explainability, clinical alignment, and generalization across diverse clinical settings. This survey aims to provide a structured roadmap for advancing AI-driven EHR modeling and clinical decision support.
For a comprehensive list of EHR-related methods, kindly refer to \url{https://survey-on-tabular-data.github.io/}.
\end{abstract}




\maketitle
\input{Sections/introduction}
\input{Sections/preliminary}
\input{Sections/Methodology/Data_Centric_Approaches}
\input{Sections/Methodology/Neural_Architectural_Approaches}
\input{Sections/Methodology/Learning_focused}

\input{Sections/Methodology/downstream_tasks}
\input{Sections/Experimental_design/exp}

\input{Sections/future_works}

\bibliographystyle{ACM-Reference-Format}
\bibliography{reference}
\end{document}

%% file: Sections/introduction.tex
\section{Introduction}
Electronic Health Records (EHRs) serve as comprehensive, longitudinal repositories of patient health data, encompassing heterogeneous modalities such as demographics, laboratory results, diagnoses, procedures, treatments, and clinical notes. As illustrated in Figure~\ref{intro::ehr_dim}, these data can be systematically characterized along three fundamental pairwise dimensions: time, feature, and value. Each core component of EHRs aligns naturally with one of these dimension pairs:
\input{Figures/intro/EHR}
(1) \textbf{Temporal clinical workflows} correspond to the time $\times$ feature dimension and capture the timing and sequence of clinical events, reflecting both the patient’s care trajectory and the physician’s decision-making process. This includes events such as symptom onset, diagnostic testing, and therapeutic interventions, and emphasizes when and why specific clinical variables are observed or recorded. 
(2) \textbf{Structured quantitative observations}, such as laboratory measurements and vital signs, correspond to the $ \text{feature} \times \text{value} $ dimension, capturing the relationship between clinical features and their measured values at specific time points;
(3) \textbf{The temporal evolution of measurements}, including trends in biomarkers or vital signs over time, reflects the time $\times$ value dimension and characterizes how clinical values change longitudinally.

Early approaches to EHRs modeling were heavily grounded in
feature engineering \cite{lipton2015learning,choi2016retain}, where researchers or engineers used their domain knowledge to define and extract salient features from raw data and provide models with the appropriate inductive bias to learn
from this limited data \cite{baytas2017patient}. 
With the advent of neural network models, salient features were
learned jointly with the training of the model itself , and hence focus shifted to architecture
engineering \cite{luo2020hitanet,li2020bERHst,li2020deepalerts}, where inductive bias was rather provided through the design of a suitable network
\cite{luo2020hitanet,huang2024dna,chang2025focus,vaswani2017attention}. 
Broadly, these architecture-based methods aim to capture temporal dependencies and structural relationships of EHRs.
For example, deep models such as transformers \cite{shi2023modelling} have been employed to model temporal dynamics in patient trajectories, enabling end-to-end learning over sequential data. 
Beyond temporal modeling, other architectures are designed to reflect the structural and relational nature of clinical data.
Tree-based models typically leverage the hierarchical structure of clinical concepts such as diagnostic categories \cite{murris2024tree,chen2024tree} or procedural taxonomies \cite{lu2023towards,pokharel2020representing}. By aligning model structure with the semantic organization of medical knowledge, these approaches incorporate meaningful inductive bias that supports clinically relevant reasoning and decision-making.
Graph-based models represent patients \cite{wu2021leveraging,lu2021collaborative,park2022graph,lu2022context,li2025time}, visits \cite{choi2020learning,zhu2021variationally}, or medical codes \cite{choi2017gram,li2022graph,shang2019pre} as nodes in a graph, where edges capture relationships such as co-occurrence, temporal succession, or connections derived from medical ontologies.

From 2020 to 2023, a pivotal shift occurred with the adoption of pretraining-based models for EHRs modeling ~\cite{zhang2023biomedclip,lai2023faithful,soenksen2022integrated}. Early pretrained models such as BEHRT \cite{li2020behrt}, MedBERT \cite{rasmy2021med}, and ClinicalBERT \cite{huang2019clinicalbert} were built on Transformer architectures and trained using domain-specific clinical data. These models used self-supervised objectives, including masked token prediction \cite{wu2024switchtab, yin2024mcm} and contrastive learning \cite{tonekaboni2021unsupervised,yeche2021neighborhood,wang2023contrast} to capture semantic, temporal, and hierarchical patterns in patient records \cite{zhang2024tacco, lee2020temporal}. 
This transition marked a shift in research focus from architecture design toward objective engineering, emphasizing the development of effective pretraining objectives and task adaptation strategies \cite{xue2023assisting, yang2023manydg}.

Recent advances in natural language processing have led to the development of large-scale foundation models for clinical applications. Models such as GatorTron \cite{yang2022gatortron}, ClinicalT5 \cite{lu2022clinicalt5}, and MedPaLM \cite{tu2024towards}, with hundreds of millions to billions of parameters, are pretrained on diverse clinical corpora including EHRs, clinical notes, and biomedical literature. These models support a broad range of clinical tasks through prompting \cite{xu2025dearllm,han2024featllm,hollmann2023large}, advancing EHRs modeling toward general-purpose clinical reasoning.
While prompting offers flexible task adaptation, fine-tuning remains essential for improving task-specific performance and domain alignment, as shown by models like 
DiaLLMs \cite{ren2025diallms},
MedFound \cite{liu2025generalist}, Me-LLaMA \cite{xie2025medical}.
To enhance factual consistency and contextual relevance, retrieval-augmented generation incorporates external knowledge, such as EHRs data, clinical guidelines, or biomedical literature \cite{chen2024tablerag, liu2022retrieve,kresevic2024optimization,wang2025colacare,wang2025medagent}. Building on these capabilities, agent-based frameworks \cite{fallahpour2025medrax} have emerged to decompose complex clinical workflows into modular subtasks, enabling interpretable, multi-step reasoning grounded in both patient data and medical knowledge \cite{pellegrini2025ehrs, mao2025ct}.

\noindent \subsection{Structure of the Survey}
This survey aims to systematically organize current knowledge in the rapidly evolving field of EHR modeling.
Section~\ref{intro::definition} introduces the problem definition, fundamental properties of EHR data, and foundational neural architectures.
Section~\ref{data_centric} presents data-centric approaches that improve data quality and quantity.
Section~\ref{neural_model} focuses on neural architecture design, covering feature-aware modules, structure-aware architectures, and temporal dependency modeling.
Section~\ref{objective} surveys learning objectives, including supervised, self-supervised, and contrastive learning methods tailored for EHR data.
Section~\ref{sec::llms} provides a comprehensive overview of recent developments in large language models (LLMs) for EHR modeling, including prompting strategies, instruction tuning, retrieval-augmented generation (RAG), and agent-based frameworks. Section~\ref{sec::multi-modality} discusses recent advances in medical vision-language models.
Section~\ref{sec::application} categorizes key downstream applications, including clinical document understanding, clinical reasoning and decision support, clinical operations Support.
Section~\ref{sec::dataset_evaluation} summarizes publicly available datasets and evaluation metrics to promote reproducibility and accessibility for new researchers.
Finally, we discusses current challenges and outlines promising directions for future research in Section~\ref{sec::challenge_future}.

To help readers better understand the structure and key concepts covered in this survey, we provide a set of supporting materials both within the paper and through a companion website. These include:

\begin{itemize}
\item A companion website\footnote{\url{https://survey-on-tabular-data.github.io/}} provides frequent updates to the survey and additional resources related to EHR modeling.
\item Table~\ref{tab:survey_llm_comparison}: Comparison with related surveys on EHR modeling.
\item 
Table~\ref{tab:design_taxonomy}: A comprehensive taxonomy of EHR modeling methods, covering training data strategies, neural architectures, learning objectives, general LLM paradigms and multimodal clinical LLMs.
\item Table~\ref{tab::agents}: A categorized collection of commonly used clinical agent workflows.
\item Figure~\ref{app:workflow}: Visualization of key clinical decision support tasks organized across the care timeline, spanning pre-encounter, encounter, and post-encounter stages.
\item Table \ref{tab::function_taxonomy} summarizes the categorization of representative clinical tasks across the care timeline. 
\item Table \ref{tab:dataset_stats} shows statistical summary of clinical datasets, including patient counts, disease coverage, data sources, and downstream tasks.
\item Table \ref{tab:dataset_comparison} shows comparison of clinical Datasets by Format, Modalities, and EHR Features

\item Table \ref{tab:metric} shows metrics on different downstream tasks.
\end{itemize}
\input{Tables/intro/related_survey}
\subsection{Related Surveys}
Based on the background discussed above, designing state-of-the-art methods for EHRs modeling involves four fundamental components: training data, model architecture, learning objectives, and foundation model adaptation strategies.
To improve both the quantity and quality of training data, data-centric techniques such as augmentation, synthesis, and harmonization have been widely adopted. To better leverage the unique properties of EHRs, neural architectures are tailored to capture: (1) heterogeneous clinical features across modalities (e.g., diagnoses, medications, lab results), (2) conditional dependencies across clinical events, and (3) the dynamic progression of patient trajectories over time. Learning objectives further guide model training through supervised, self-supervised, and contrastive paradigms, facilitating generalization under label-scarce conditions.
With the rise of LLMs, EHRs modeling increasingly incorporates foundation model adaptation through prompting, instruction tuning, fine-tuning, and retrieval augmentation. These strategies enable the reuse of pretrained clinical or general-purpose models for a broad range of downstream tasks, improving flexibility, generalization, and interpretability in clinical settings.

Recent surveys on deep learning for EHRs have largely concentrated on isolated components of the modeling pipeline. Xie et al. \cite{xie2022deep} and Li et al. \cite{li2022neural} focus on architectural design, offering limited treatment of data-centric strategies or language model integration. Wornow et al. \cite{wornow2023shaky} evaluate clinical language models but do not address broader modeling frameworks. Du et al. \cite{du2024generative} concentrate on generation tasks, while Liu et al. \cite{liu2024survey} comprehensively review clinical language models, focusing on data sources and downstream applications but overlooking core modeling methodologies. Wang et al. \cite{wang2025survey} explore clinical agents from a system-level perspective, without connecting to foundational modeling practices. In contrast, our survey presents a unified and comprehensive view encompassing data-centric approaches, neural architecture design, learning objectives, and contemporary language model paradigms. This includes prompting, fine-tuning, retrieval-augmented generation, and the integration of agent-based frameworks for clinical decision-making.
Table \ref{tab:survey_llm_comparison} highlights the distinctions between our survey and existing literature.

\subsection{Survey Scope and Literature Collection}
For literature review, we use the following keywords and inclusion criteria to collect literatures.

\noindent \textbf{Keywords}. ``EHRs'' or ``EMRs'' or ``Biomedical Data'' or ``Tabular Data'' or ``Clinical Large Language Models''.
We use these keywords to search well-known repositories, including ACM Digital Library, IEEE Xplore, Google Scholars, Semantic Scholars, and DBLP, for the relevant papers.

\noindent \textbf{Inclusion Criteria}
Related literatures found by the above keywords are further filtered by the following criterion. Only papers meeting these criteria are included for review.
\begin{itemize}
\item Written exclusively in the English language
\item Focused on approaches based on machine learning, deep learning, neural networks or LLMs.
\item Published in or after 2020 in reputable conferences or high-impact journals
\end{itemize}


\subsection{Contributions}
This survey aims to systematically organize the evolving landscape of EHRs modeling, spanning foundational deep learning approaches to emerging paradigms based on LLMs. To the best of our knowledge, this is the first comprehensive survey that unifies data-centric strategies, neural architecture design, learning objectives, and LLM-based reasoning within a single framework for EHRs applications.

We introduce a novel taxonomy that categorizes EHRs modeling methods across four key design dimensions: (1) training data quantity and quality, (2) neural architectures tailored to the heterogeneity and temporal nature of EHRs, (3) learning objectives including supervised, self-supervised etc, and (4) LLM paradigms such as prompting, fine-tuning, retrieval-augmented generation, and agent-based reasoning. Table~\ref{tab:survey_llm_comparison} presents a comparative overview of existing surveys along these dimensions.
Our main contributions are as follows:

\begin{itemize}[leftmargin=*, itemsep=1pt, topsep=2pt]
\item \textbf{Comprehensive scope}: We conduct an extensive review of over *** papers, integrating recent advances in EHRs modeling from foundational neural methods to cutting-edge LLM-driven frameworks.
    
    \item \textbf{Unified taxonomy}: We propose a principled and extensible taxonomy that spans data-centric methods, architectural innovations, learning objectives, and LLM-based strategies, offering a structured lens for understanding model design.
    \item \textbf{Coverage of emerging trends}: We highlight the emergence of pretraining, prompting, instruction tuning, retrieval-augmented generation, and agent-based clinical systems, offering timely insights into the evolving role of LLMs in clinical reasoning and decision support.
    \item \textbf{Resource and evaluation guide}: We summarize key benchmarks, datasets, and evaluation protocols to promote standardized evaluation and identify open challenges and future directions in EHRs modeling.
\end{itemize}

%% file: Figures/intro/EHR.tex
\begin{wrapfigure}{r}{0.49\textwidth}  
  \centering
  \vspace{-11pt} \includegraphics[width=0.45\textwidth]{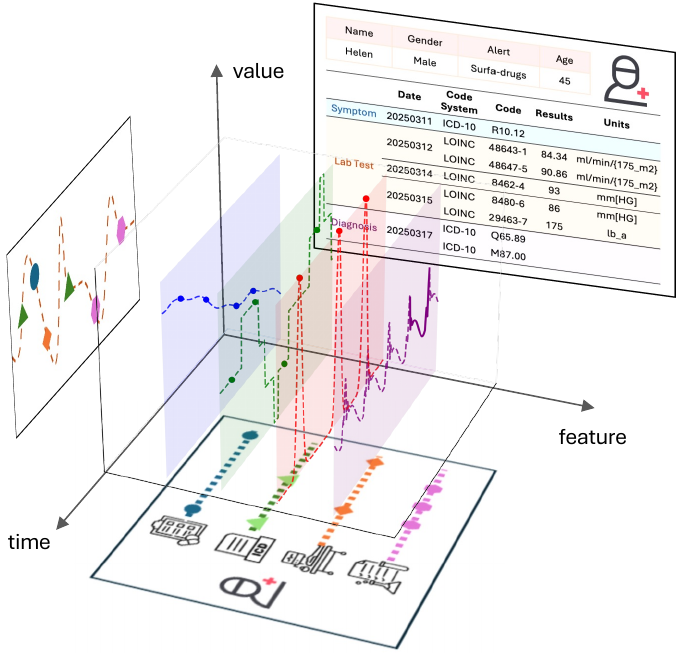}
  \caption{An example of EHRs.EHRs can be viewed through three projections: (1) $ \text{time} \times \text{feature} $, representing clinical workflows and care trajectories; (2) $ \text{feature} \times \text{value} $, capturing structured quantitative observations such as lab results; and (3) $ \text{time} \times \text{value} $, describing the temporal evolution of clinical measurements.}
  \label{intro::ehr_dim}
  \vspace{-8pt} 
\end{wrapfigure}

%% file: Tables/intro/related_survey.tex
\begin{table*}[ht]
\centering
\resizebox{1\textwidth}{!}{%
\small
\begin{tabular}{
l|c|   
p{0.9cm} p{0.9cm}| 
p{1.2cm}|          
p{2cm}|            
p{1cm} p{1cm} p{1cm} p{1cm}   
}
\toprule
\textbf{Survey} & \textbf{Time} &
\multicolumn{2}{c|}{\textbf{Data‐Centric}} & 
\textbf{Arch.\ Focus} & \textbf{Learning Objective} &
\textbf{Fine‐Tune} & \textbf{Prompt} & \textbf{RAG} & \textbf{Agent} \\
\cmidrule{3-4}
& & Gen. & Aug. & & & & & & \\
\midrule
Xie Feng et al.~\cite{xie2022deep}         & 2022 & \xmark & \xmark & \cmark & \xmark & \xmark & \xmark & \xmark & \xmark \\
Li Irene et al.~\cite{li2022neural}        & 2022 & \cmark & \xmark & \cmark & \xmark & \xmark & \xmark & \xmark & \xmark \\
Wornow Michael et al.~\cite{wornow2023shaky} & 2023 & \xmark & \xmark & \xmark & \xmark & \cmark & \xmark & \xmark & \xmark \\
Du Xinsong et al.~\cite{du2024generative}  & 2024 & \cmark & \cmark & \xmark & \xmark & \xmark & \xmark & \xmark & \xmark \\
Liu Lei et al.~\cite{liu2024survey}        & 2024 & \xmark & \xmark & \xmark & \xmark & \cmark & \cmark & \cmark & \cmark \\
Wang Wenxuan et al.~\cite{wang2025survey}  & 2025 & \xmark & \xmark & \xmark & \xmark & \xmark & \xmark & \xmark & \cmark \\
\textbf{Our Survey}                        & 2025 & \cmark & \cmark & \cmark & \cmark & \cmark & \cmark & \cmark & \cmark \\
\bottomrule
\end{tabular}
}
\caption{Comparison of survey scope across training‐data strategies, neural architectures, learning objectives, and LLM paradigms.  
\textbf{Gen.}: Synthetic data generation; 
\textbf{Aug.}: Data augmentation; 
\textbf{Arch.}: Neural Architecture.}

\label{tab:survey_llm_comparison}
\end{table*}

%% file: Sections/preliminary.tex
\section{Preliminary}
\label{intro::definition}
This section provides the definitions and notations used in the paper, describes downstream tasks for tabular data analysis, and highlights the unique properties of tabular data.

\subsection{Definitions}
\begin{definition}
\textbf{(EHRs).} Electronic Health Records (EHRs) are digitized collections of patient health information that include static features (e.g., demographics) and longitudinal time-stamped observations (e.g., lab results, vital signs), enabling the analysis of both cross-sectional and temporal aspects of patient care.  
Denote $D$ as an EHR dataset with $N$ patients, where $X$ represents patient data and each patient's record is given by 
\[
X_i = \big(\boldsymbol{x}^{\text{static}}_i, \{ (\boldsymbol{t}_{ij}, \boldsymbol{x}_{ij}) \}_{j=1}^{T_i} \big),
\]
where $\boldsymbol{x}^{\text{static}}_i$ denotes static features and $\{ (\boldsymbol{t}_{ij}, \boldsymbol{x}_{ij}) \}$ represents the sequence of time-stamped observations for patient $i$.
\end{definition}

\begin{definition}
\textbf{(Classification).}
EHR classification aims to assign predefined clinical labels 
$Y = \{y_1, y_2,\ldots,y_C \}$ to each patient record based on their health data, including static features and longitudinal time-stamped observations.  
For binary classification, $y_i \in \{-1,1\}$; for multi-class classification, $y_i \in \{1,2,\ldots,C\}$.
\end{definition}

\begin{definition}
\textbf{(Regression).}
EHR regression shares the same setup as classification but differs in the target type: the goal is to predict a continuous value $y \in \mathbb{R}$ for each patient.
\end{definition}

\begin{definition}
\textbf{(Clustering).}  
EHR clustering aims to partition patients into a set of clusters $G = \{g_1, g_2,\ldots, g_K\}$ by maximizing similarity among patients within the same cluster while maximizing dissimilarity across different clusters.
\end{definition}

\begin{definition}
\textbf{(Temporal Forecasting).}  
EHR temporal forecasting aims to predict future clinical measurements or events at one or multiple future time points, given a patient's historical static features and longitudinal time-stamped observations.  
The goal is to model temporal dependencies to estimate outcomes beyond the observed time horizon.
\end{definition}

\begin{definition}
\textbf{(Imputation of Missing Values).}  
EHR imputation aims to fill missing values with plausible estimates to enable subsequent analysis.  
Given EHR data $X$ and a known binary mask matrix $M \in \{0,1\}^{|X|}$, an entry $x_i$ is missing if $m_i = 0$ and observed otherwise.  
The imputed data is computed as:
\[
X_{\text{imputed}} = X \odot M + \hat{X} \odot (1 - M),
\]
where $\odot$ denotes elementwise multiplication.
\end{definition}

\begin{definition}
\textbf{(Survival Prediction).}
EHR survival prediction aims to model the time until an event of interest (e.g., death, readmission) occurs for each patient, using both static features and longitudinal time-stamped observations as inputs. More specifically, survival prediction aims to predict $(T_i, \delta_i)$, with $T_i$ representing the observed time (either event time or censoring time) and $\delta_i \in \{0,1\}$ indicating whether the event was observed or censored.
\end{definition}

\begin{definition}
\textbf{(Retrieval).}  
EHR retrieval aims to identify patients most similar to a given query.  
Given a query patient $X$ and a similarity measure $f(\cdot)$, the goal is to find an ordered list 
\[
Q = \{X_i\}_{i=1}^k
\]
containing the $k$ most similar patients in the dataset.
\end{definition}

\begin{definition}
\textbf{(Synthetic Data Generation).}
EHR synthetic data generation aims to learn the underlying distribution of real patient records and produce realistic, privacy-preserving synthetic EHR samples.  
Given an original dataset $D$, a generative model outputs new samples $X'$ that maintain statistical properties of $D$ while protecting patient confidentiality.
\end{definition}

\subsection{Foundational Properties of EHRs}

\subsubsection{Heterogeneity}
EHR data exhibit significant heterogeneity, encompassing categorical variables (e.g., diagnostic codes), binary indicators (e.g., symptom presence), continuous measurements (e.g., laboratory values, vital signs), and unstructured textual notes.  
This intrinsic heterogeneity poses modeling challenges, requiring specialized encoding strategies and architectures to capture diverse semantics and statistical patterns.

\subsubsection{Temporal Dependency}
EHR data show strong temporal dependencies due to the sequential nature of clinical events, where diagnoses, treatments, and observations occur over time, with past events influencing future outcomes.  
We categorize temporal dependencies into three dimensions:
\begin{enumerate}
\item \textbf{Irregular/Asynchronous Sampling}: Clinical observations are recorded at irregular, asynchronous intervals across modalities. Variability in time gaps and event timing challenges traditional time-series models, requiring time-aware mechanisms such as decay functions, interpolation, or continuous-time models.
\item \textbf{Multi-Timescale Dynamics}: EHR data contain signals evolving at multiple temporal resolutions, such as short-term physiological changes alongside long-term chronic conditions. Effective models must integrate across timescales, often via hierarchical or multi-resolution designs.
\item \textbf{Conditional Clinical Sequences}: EHRs reflect structured, condition-dependent sequences, where each step depends on prior outcomes. For example, diagnostic tests may precede confirmatory procedures and treatments. Modeling these sequences requires architectures that capture event order and conditional dependencies for accurate prediction and individualized reasoning.
\end{enumerate}

\subsubsection{Semantic Ambiguity}
EHRs often contain semantically ambiguous entries. The same clinical code, term, or lab value can signify different meanings depending on context such as the patient’s demographics, comorbidities, or clinical setting. For instance, a fasting blood glucose level of 110 mg/dL is considered normal in most healthy adults but may be classified as abnormal in a patient with a history of gestational diabetes or in children at risk of type 1 diabetes.
This semantic ambiguity complicates data annotation, limits model generalization, and poses challenges for clinical decision support. Addressing this requires context-aware representation learning and integration of temporal, demographic, and knowledge-based cues.

\subsubsection{Domain-Specificity}
EHRs encode rich, domain-specific information that is deeply intertwined with medical knowledge. This domain specificity manifests in at least three critical ways.  
(1) Clinical codes whose meanings are contextually defined, such as ICD-10 code I50.9 indicating unspecified heart failure;
(2) Measurement interpretations. For example, a creatinine level of 1.3 mg/dL is considered normal for a healthy adult;
(3) Decision-support evidence embedded in practice guidelines, such as initiating insulin therapy when A1C exceeds $9\%$, or withholding anticoagulants in patients with high bleeding risk despite atrial fibrillation.

\subsubsection{Sparsity} 
EHR data are inherently sparse, with each patient record capturing only a limited subset of the vast clinical vocabulary. This sparsity stems from multiple factors: (1) irregular sampling due to heterogeneous visit patterns, (2) variability in physician documentation style and diagnostic certainty, and (3) data omissions or recording errors. Notably, the missingness is often non-random, reflecting underlying clinical workflows and decision-making rather than absence of clinical relevance.
\subsubsection{Class Imbalance}  
EHR prediction tasks often exhibit significant class imbalance, where clinically important outcomes such as disease onset, adverse events, or mortality occur much less frequently than the majority class. This uneven distribution challenges model training by biasing predictions toward more common outcomes and reducing the model's ability to detect rare but critical events. Additionally, the extent of imbalance can vary across patient subgroups, raising concerns about fairness and equitable model performance.

\subsection{Basic Neural Architectures}
This section outlines key neural architectures frequently used in healthcare AI, emphasizing their core components and operational principles.

\subsubsection{Multi-Layer Perceptron (MLP)}
A Multi-Layer Perceptron (MLP) is a type of feedforward neural network composed of multiple stacked layers, where each layer applies an affine transformation followed by a nonlinear activation function.  
MLPs are fully connected architectures, meaning every neuron in a given layer is connected to all neurons in the previous layer, allowing them to learn complex, nonlinear relationships among input features.  
They lack inherent spatial or temporal inductive biases, making them especially well-suited for structured tabular data where features do not exhibit spatial locality or sequential dependencies.  
Their flexibility and simplicity have made MLPs a common baseline for tabular data modeling, capable of capturing interactions among heterogeneous feature types in electronic health records (EHRs) and other structured datasets.

\subsubsection{Convolutional Neural Network (CNN)}
A Convolutional Neural Network (CNN) is a deep neural architecture designed to learn hierarchical spatial representations by leveraging local connectivity and parameter sharing.  
Each convolutional layer applies a set of learnable filters to small, localized regions of the input, enabling the model to detect spatially local patterns that can generalize across the entire input space.  
CNNs are typically composed of multiple stacked convolutional layers interleaved with nonlinear activation functions (e.g., ReLU), pooling layers (e.g., max or average pooling) to reduce spatial resolution and promote translational invariance, and normalization layers (e.g., batch normalization) to stabilize and accelerate training.  
As the network depth increases, these layers progressively capture more abstract and semantically meaningful features.
In medical imaging, CNNs are widely used to analyze modalities such as X-rays, CT scans, and MRIs.  
Their ability to capture local anatomical structures and spatial hierarchies makes them highly effective for tasks including segmentation, classification, and the detection of pathological patterns.

\subsubsection{Transformer}
Transformers are deep neural architectures designed to model global dependencies in data using self-attention mechanisms.   
Transformers employ multi-head self-attention and positional encoding to learn diverse representation subspaces simultaneously, enhancing the model's ability to capture complex patterns. 
The architecture typically consists of stacked layers that include multi-head self-attention modules, position-wise feedforward networks, residual connections, and normalization layers, allowing deep, stable, and highly expressive models.
Originally developed for natural language processing (NLP) tasks such as machine translation and text generation, Transformers have achieved state-of-the-art performance across numerous domains.  
They are increasingly adopted in multimodal learning scenarios, including applications that integrate text, images, and structured clinical data.

\subsubsection{Capsule Network}
Capsule Networks are neural architectures that use vector-valued units, known as \emph{capsules}, to represent both the probability of a feature's presence and its properties such as pose, orientation, and scale.  
Each capsule outputs a vector whose length indicates the likelihood of the feature, while its orientation encodes detailed instantiation parameters.
Connections between capsule layers are learned through transformation matrices that map lower-level capsule outputs to predictions for higher-level capsules.  
A key mechanism called dynamic routing iteratively determines coupling coefficients that control how strongly lower-level capsules contribute to higher-level representations, adjusting these weights based on the level of agreement among predictions.  
This approach enables the network to model part–whole relationships explicitly, supporting viewpoint-invariant recognition and improving interpretability in imaging tasks by preserving spatial and compositional information.

\subsubsection{Neural Logic Network (NLN)}
Neural Logic Networks (NLNs) are neural architectures designed to approximate logical operations such as AND, OR, and NOT in a differentiable manner.  
They bridge the gap between symbolic logic and continuous optimization by representing Boolean functions as smooth, learnable transformations.  
Instead of discrete logical values, NLNs operate on continuous inputs within the $[0,1]$ interval, enabling gradient-based learning of logical rules through parameterized neural layers.  
Each NLN layer learns to approximate desired logical operations by applying nonlinear activation functions to affine transformations of its inputs.  
By stacking such layers, NLNs can capture complex, structured logical dependencies across multiple features.  
This approach supports end-to-end differentiable learning of rule-based reasoning, making NLNs well-suited for clinical decision-support tasks where interpretability, transparency, and the ability to encode domain knowledge are critical.

\subsubsection{Large Language Models (LLMs)}
LLMs are deep neural architectures based on the Transformer framework, pretrained on large text corpora using language modeling objectives such as next-token prediction.  
They typically consist of stacked Transformer blocks with multi-head self-attention, position-wise feedforward layers, token and positional embeddings, and a language modeling head for output generation.  
LLMs can be extended with adapters or additional modules to support multimodal inputs, enabling integration of text with other data types such as images.

%% file: Sections/Methodology/Data_Centric_Approaches.tex
\input{Tables/paper_list}

\section{Data-Centric Approaches}
\label{data_centric}
\input{Figures/section3/figure_section3}
This section reviews methods that aim to enhance model performance by improving the quality and utility of the training data itself, rather than modifying model architectures or learning objectives. These data-centric approaches focus on transforming or augmenting the input data to better expose meaningful patterns, temporal trends, and clinically relevant features within EHRs. As illustrated in Fig.\ref{fig:data-centric}, we categorize these methods into two main objectives: (1) improving data quality and (2) increasing data quantity.
\subsection{Improving Quality}
\subsubsection{Sample Selection}
Recent advances in data-centric EHR research emphasize the importance of selecting high-quality and informative samples to enhance model performance, reduce annotation costs, and improve robustness.
Existing works can be categorized into three main directions: (1) \textbf{Active learning}, (2) \textbf{Influence-based methods}, and (3) \textbf{Statistical sampling approaches}.
\underline{\textit{Active learning}} prioritizes samples that maximize model uncertainty or learning gain. For example, Ferreira et al. \cite{ferreira2021active} used uncertainty sampling to reduce labeled data requirements for clinical note classification. 
\underline{\textit{Influence-based methods}} estimate the impact of individual samples on model performance. In2Core \cite{joaquin2024in2core} selects core subsets via influence functions to preserve task-relevant information. Yang \cite{yang2024addressing} prunes harmful samples to mitigate label noise.
\underline{\textit{Statistical sampling approaches}} optimize validation set construction. Shepherd \cite{shepherd2023multiwave} proposed a multi-wave sampling framework guided by influence measures to improve the efficiency of clinical chart reviews.

\subsubsection{Input Space Transformation}
To improve model robustness, generalizability, and interpretability in clinical prediction tasks, recent research has explored transforming or augmenting the input space into enriched representations. Existing approaches can be broadly classified into two categories:
(1) \textbf{Code perturbation}, and
(2) \textbf{Semantic augmentation}.
\underline{\textit{Code perturbation}} applies controlled modifications to input sequences, such as shuffling medical codes within or across visits, in order to promote temporal invariance and reduce overfitting. For example, Choi $\&$ Kim et al. \cite{choi2024data} proposed dynamic token masking and reordering to regularize sequential learning on longitudinal health records.
\underline{\textit{Semantic augmentation}} incorporates domain knowledge, such as hierarchical code groupings or clinical ontologies, to enrich input representations. For example, Bloore et al. \cite{bloore2022semantic} introduced H-BERT, which semantically decomposes hierarchical medical codes into graph-aware embeddings to improve representation learning.
More recently, Zhou et al. \cite{zhou2025generating} proposed HiSGT, which integrates hierarchical graph structures and semantic embeddings (e.g., from ClinicalBERT) to reduce input sparsity and capture richer clinical context.
DiaLLMs ~\cite{ren2025diallms} annotate the clinical meaning of symptoms and diagnostic codes to enhance understanding and reasoning in clinical language models.
RAM‑EHR \cite{xu2024ram} uses a dense retrieval pipeline to fetch relevant text from multiple knowledge sources (e.g., clinical notes). 

\subsubsection{Multi-source Learning}
EHRs are inherently decentralized and sparse, often collected across different institutions, devices, and settings. To address data incompleteness and improve model generalizability, recent studies explore multi-source representation learning by integrating heterogeneous data sources and leveraging cross-institutional knowledge.
UNITE~\cite{chen2021unite} combines structured EHR data, patient demographics, and public web-based health information using an adaptive deep kernel and stochastic variational inference. This approach enhances health risk prediction while providing well-calibrated uncertainty estimates in high-dimensional, multi-source environments.
Chaudhari et al.~\cite{chaudhari2021lowcount} evaluate the generalizability of a deep learning–based image quality enhancement model across multiple institutions and scanner types in a reduced-count PET imaging setting, highlighting the importance of multi-site validation.
Sadilek et al.~\cite{sadilek2021privacy} apply federated learning with differential privacy to enable collaborative clinical and epidemiological modeling across diverse units, diseases, and tasks without compromising data confidentiality.
MUGS~\cite{li2024multisource} presents a transfer learning framework that fuses graphs from pediatric and general EHRs with medical ontologies, generating adaptive embeddings for robust patient profiling. It outperforms existing baselines on pediatric-specific tasks such as pulmonary hypertension detection.
MH-pFLID~\cite{xie2024mh} introduces a lightweight personalized federated learning framework using a messenger-based architecture with transmitter and receiver modules. 

\subsubsection{Learning with Knowledge Graphs}
GRAPHCARE \cite{jiang2025reasoning} generates personalized knowledge graphs for patients by fusing EHR data with external biomedical KGs and LLM-derived insights. These graphs are then modeled using a novel Bi-attention Augmented Graph Neural Network (BAT-GNN) to improve clinical prediction tasks.
KARE \cite{jiang2023graphcare} enhances clinical predictions by integrating multi-source medical knowledge (biomedical databases, literature, LLMs) into a hierarchically structured KG, leveraging community-aware retrieval and LLM reasoning to generate interpretable, context-enriched patient data. 
MedRAG \cite{zhao2025medrag} integrates a four-tier hierarchical diagnostic knowledge graph with EHR retrieval to enhance medical RAG frameworks. It dynamically combines KG-derived disease differentiators with similar patient records and uses LLM reasoning to generate precise diagnostic/treatment recommendations while proactively suggesting follow-up questions for clinical decision support.
EHR-QA \cite{park2021kgqa} proposes a graph-based  framework where natural language queries are translated to SPARQL queries, arguing graph structures eliminate relational JOIN operations and more naturally encode clinical entity relationships.
\cite{lu2021readmission} proposes representing EHR discharge summaries as knowledge-augmented multiview graphs and applies GCNs for ICU readmission risk prediction, enabling clinically interpretable text-driven assessment.
SeqCare \cite{seqcare2023} proposes a sequential framework for diagnosis prediction that first applies graph contrastive learning to reduce bias in medical KG representations, then jointly refines KG supervision signals and graph structure to minimize uncertainty, and finally employs self-distillation in the optimized latent space to filter irrelevant features.
KerPrint \cite{yang2023kerprint} introduces a KG-enhanced approach using time-aware KG attention to mitigate knowledge decay for trustworthy retrospective interpretations. It also incorporates an element-wise attention for global knowledge selection from local KG representations to enable prospective clinical inference.
TabR \cite{gorishniy2024tabr} retrieves similar instances using an attention-based k-NN module embedded in a feed-forward network
MedPath \cite{medpath2021} augments clinical risk prediction by extracting personalized knowledge graphs (PKGs) of disease progression paths from medical KGs, encodes them via multi-hop GNN message passing from symptoms to diseases, and provides interpretable predictions by tracing symptom-to-disease pathways.

\subsection{Improving Quantity}
Sample generation is a popular technique that increases
the size and diversity of the training data when the data is scarce. It explicitly creates new samples
by transformation or generative models.
\subsubsection{Generating Heterogeneous EHRs}
EHRs consist of heterogeneous data types, including static demographics, discrete codes (e.g., diagnoses, medications, procedures), continuous time-series (e.g., vital signs, lab results). Generating synthetic heterogeneous EHRs poses significant challenges due to the need to jointly model mixed-type variables while preserving inter-modality dependencies and clinical semantics.
Early methods primarily focused on generating single-modality data, such as binary diagnosis vectors \cite{baowaly2019synthesizing}, which limited their ability to capture the rich interactions inherent in real-world EHRs. Recent approaches have addressed this limitation by designing multi-branch architectures and unified latent representations that can accommodate various data modalities.
One representative line of work involves modality-specific encoders and generators that map different types of inputs into a shared latent space. For example, EHR-M-GAN \cite{li2023generating} employs dual variational autoencoders for categorical and continuous data, and couples them with LSTM-based generators to capture temporal correlations across modalities.
FLEXGEN-EHR \cite{he2024flexible} define an optimal transport module to align and accentuate the common feature space of heterogeneity of EHRs.

\subsubsection{Generating Temporal EHRs}
Generating realistic temporal EHRs presents unique challenges due to the irregularity, sparsity, and multi-scale nature of clinical events. Based on these challenges, existing methods for temporal EHR generation can be broadly categorized into three lines of work: 1) \textbf{sequential modeling}, 2) \textbf{handling irregular sampling}, and 3) \textbf{capturing long-range temporal dependencies}.
\underline{\textit{Sequential Modeling of Clinical Events}} formulates EHRs as event sequences and adopts autoregressive or sequence-to-sequence models to generate future clinical visits.
Early methods \cite{chen2024generating,guan2019method,sun2021generating} utilized RNNs to encode visit histories and capture short- to mid-range dependencies. Recent advances \cite{li2022hi,yang2023transformehr} leverage Transformer-based architectures to model long-range temporal dependencies with improved scalability. These models generate visit-level or token-level clinical events (e.g., diagnoses, medications, procedures) by learning the underlying temporal structure of patient trajectories.
\underline{\textit{Handling Irregular Sampling}} addresses the challenge that clinical data are often collected at non-uniform intervals, where the patterns of missingness reflect both clinical relevance and operational constraints. Existing approaches for modeling such data generally fall into two categories. The first is missingness-aware modeling, which explicitly encodes time gaps and observation masks using time embeddings or attention mechanisms \cite{wang2024igamt,yu2024smart,ren2022generative,he2024flexible}. The second is imputation-based modeling, which reconstructs irregular samples by interpolating or imputing missing values. For example, Moon \cite{moon2022survlatent, sheu2025continuous,sheu2025continuous,xiao2024ivp,he2023meddiff} apply Neural ODEs to learn continuous-time representations for imputation. Shukla \cite{shukla2021multi} combines hand-crafted interpolation embeddings through a gating mechanism.
\underline{\textit{Capturing long-range temporal dependencies}} is critical for modeling chronic conditions and generating clinically meaningful longitudinal records. Traditional recurrent neural networks often struggle with vanishing gradients and limited memory, motivating the development of advanced architectures to extend the temporal horizon. Existing approaches can be broadly categorized into three main strategies:
(1) Hierarchical Temporal Modeling.
These models structure clinical timelines into multiscale representations, allowing the abstraction of long-term patterns while preserving fine-grained temporal detail. For example, Hi-BEHRT \cite{li2022hi} employs hierarchical encoders with segment-level self-attention to process lengthy medical histories efficiently. HALO \cite{theodorou2023synthesize} further separates visit-level and event-level modeling, capturing cross-visit patterns while maintaining intra-visit resolution. Such designs are well-suited for synthesizing multi-year trajectories in chronic disease settings.
(2) Memory-Augmented Transformers.
To overcome the context window limitations of standard Transformers, memory-based approaches explicitly incorporate long-term storage. Foresight \cite{kraljevic2024foresight}  introduces an external memory module that retains past contextual information across visits. 
(3) Semantics-Guided Temporal Attention.
Another line of work integrates domain knowledge into temporal modeling to prioritize clinically relevant long-range dependencies. HiSGT \cite{zhou2025generating} constructs hierarchical graphs over diagnosis and procedure codes to inform temporal attention. 

%% file: Tables/paper_list.tex
\begin{table*}[htbp]
\centering
\resizebox{1.0\textwidth}{!}{ 
\begin{tabular}{ccccp{8cm}}
\toprule
\Huge \textbf{Design Element} 
& \textbf{\Huge Category} 
& \multicolumn{2}{c}{\textbf{\Huge Coarse-to-Fine Taxonomy}} 
& \textbf{\Huge Reference} 
\\
\midrule
\multirow{9}{*}{\begin{tabular}[c]{@{}l@{}} \Huge Data-Centric Approaches\\
\end{tabular}}  
& \multirow{4}{*}{\Huge Improving Quality} 
& \multirow{3}{*}{\Huge Sample Selection}
& \Huge Active learning
& \cite{ferreira2021active}
\\
&
&
& \Huge Influence-based methods
& \cite{joaquin2024in2core}, \cite{yang2024addressing}
\\
&
&
&\Huge Statistical sampling
& \cite{shepherd2023multiwave}
\\
\cmidrule(lr){4-5}
& 
& \multirow{3}{*}{\Huge Input Space Transformation}
& \Huge Code Perturbation
& \cite{choi2024data}
\\
&
&
&\Huge Semantic Augmentation
&\cite{bloore2022semantic}, \cite{zhou2025generating}, \cite{ren2025diallms}, \cite{xu2024ram}
\\
\cmidrule(lr){4-5}
&
&\Huge Multi-source Learning
&
& \cite{chaudhari2021lowcount},
\cite{li2024multisource},
\cite{sadilek2021privacy},
\cite{chen2021unite},
\cite{xie2024mh}
\\
&
&\Huge Learning with Knowledge Graphs
&
&GRAPHCARE \cite{jiang2025reasoning},   
KARE \cite{jiang2023graphcare},   
MedRAG \cite{zhao2025medrag},       
EHR-QA \cite{park2021kgqa},         
\cite{lu2021readmission},   
SeqCare \cite{seqcare2023},          
KerPrint \cite{yang2023kerprint},     
TabR \cite{gorishniy2023tabr},
MedPath \cite{medpath2021}          
\\
\cmidrule(lr){2-5}
& \multirow{5}{*}{\Huge Increasing Quantity} 
&               
& — 
& \cite{baowaly2019synthesizing}, EHR-M-GAN \cite{li2023generating}, FLEXGEN-EHR \cite{he2024flexible}
\\
&
&\Huge Generating Heterogeneous EHRs 
&
&
\\
\cmidrule(lr){4-5}
& 
& \multirow{3}{*}{\Huge Generating Temporal EHRs} 
& \Huge Sequential Modeling of Clinical Events      
& \cite{chen2024generating}, \cite{guan2019method}, \cite{sun2021generating}, \cite{li2022hi}, \cite{yang2023transformehr}
\\
& 
&                            
& \Huge Irregularly Sampled EHRs          
& \cite{wang2024igamt}, \cite{yu2024smart}, \cite{ren2022generative}, \cite{he2024flexible}, \cite{moon2022survlatent}, \cite{sheu2025continuous}, \cite{xiao2024ivp}, \cite{he2023meddiff}, \cite{shukla2021multi}
\\
& 
&                            
& \Huge Long-Range Temporal Dependencies     
& Hi-BEHRT \cite{li2022hi}, HALO \cite{theodorou2023synthesize}, Foresight \cite{kraljevic2024foresight}, HiSGT \cite{zhou2025generating}, ChronoFormer \cite{zhang2025chronoformer}
\\
\midrule
\multirow{11}{*}{\begin{tabular}[c]{@{}l@{}}\Huge Neural Modeling Design
\end{tabular}} 
& \multirow{2}{*}{\Huge Feature-Aware Modules} 
&\Huge Discretization and Binning
&
&\cite{lee2024binning}, TabLog \cite{ren2024tablog}, \cite{karimireddy2020byzantine}, \cite{gorishniy2022embeddings}, \cite{zantedeschi2021learning}
\\
&
&\Huge Kernel-Based Encoders
&
&\cite{good2023feature}
\\
\cmidrule(lr){2-5}
& \multirow{7}{*}{\Huge Structure-aware Architecture Design} 
& \multirow{3}{*}{\Huge Tree-based} 
& \Huge Tree-Guided Feature Selection             
& \cite{murris2024tree}, \cite{chen2024tree}, \cite{chen2016xgboost}, \cite{good2023feature}\\
&
&
& \Huge Tree-Guided Clinical Coding
& \cite{liu2022treeman}, \cite{lu2023towards}
\\
&
&
& \Huge Tree-Based Event Sequence Modeling
& \cite{pokharel2020representing}, \cite{song2024latent}
\\
\cmidrule(lr){3-5}
& 
& \multirow{4}{*}{\Huge Graph-based} 
& \Huge Code-Level
& GRAM \cite{choi2017gram}, G-BERT \cite{shang2019pre}, \cite{li2022graph} \\
& 
& 
& \Huge Visit-Level
& GCT \cite{choi2020learning}, \cite{zhu2021variationally} \\
& 
& 
& \Huge Patient-Level
& ME2Vec \cite{wu2021leveraging}, \cite{lu2021collaborative}, \cite{park2022graph}, \cite{lu2022context}, \cite{li2025time}, \cite{kim2025hi}, funGCN \cite{boschi2024functional} \\
& 
& 
& \Huge Hybrid-Level
& \cite{lee2020harmonized}, \cite{poulain2024graph} \\
\cmidrule(lr){3-5}
& 
& \Huge Rule-based Models 
&                 
& \cite{lowe2022logical}, SQRL \cite{naik2023machine}, TabLog \cite{ren2024tablog}, ConSequence \cite{theodorou2024consequence}
\\
&
&\Huge Neural Additive Model
&
& NAMs \cite{agarwal2021neural}, \cite{cheng2024arithmetic}, SIAN \cite{enouen2022sparse}, \cite{ibrahim2023grand}, \cite{lyu2023towards}
\\
& 
& \Huge Hierarchical and Structured Transformers
&          
& Transformer \cite{vaswani2017attention}, \cite{wang2024transformer}, HiTANet \cite{luo2020hitanet}, RAPT \cite{ren2021rapt}, TransformEHR \cite{yang2023transformehr}, BEHRT \cite{li2020behrt}, MedBERT \cite{rasmy2021med}, RareBERT \cite{prakash2021rarebert}, CEHR-BERT \cite{poulain2022few}, BEHRT-CVD \cite{poulain2021transformer}, DL-BERT \cite{chen2022dl}, GT-BEHRT \cite{poulain2024graph}
\\
\cmidrule(lr){2-5}
& \multirow{4}{*}{\Huge Temporal Modeling Design} 
& \multirow{2}{*}{\Huge Irregular/Asynchronous Sampling} 
& \Huge Implicit Temporal Handling
& GRU-D \cite{che2018recurrent}, Neural ODEs \cite{chen2018neural}, SeFT \cite{horn2020set}, ContiFormer \cite{chen2023contiformer}, DNA-T \cite{huang2024dna}, \cite{chung2020deep}, \cite{zhang2021graph}, \cite{yin2024sepsiscalc}, \cite{chen2024predictive}, Warpformer \cite{zhang2023warpformer}, \cite{li2023time}, \cite{chen2025integrating}
\\
&
&
& \Huge Explicit Temporal Alignment
& \cite{katsuki2022cumulative}, \cite{li2021imputation}, \cite{shukla2019interpolation}, \cite{shukla2021multi}, \cite{zhang2023improving}, \cite{zhong2024hybrid}, \cite{tipirneni2022self}
\\
\cmidrule(lr){4-5}
& 
&\Huge Multi-Timescale Dynamics
&
& Autoformer \cite{wu2021autoformer}, FedFormer \cite{zhou2022fedformer}, StageNet \cite{gao2020stagenet}, DeepAlerts \cite{li2020deepalerts}, AdaCare \cite{ma2020adacare}, \cite{nguyen2021clinical}, \cite{ma2020adacare}, \cite{shi2023modelling}, \cite{li2020deepalerts}\\
&
&\Huge Conditional Clinical Sequences 
&
& \cite{placido2023deep}, \cite{chen2021novel}, DeepAlerts \cite{li2020deepalerts}, Diaformer \cite{chen2022diaformer}, MICRON \cite{yang2021change}, ConCare \cite{ma2020concare}, DiaLLMs \cite{ren2025diallms}
\\
\midrule
\multirow{5}{*}{\Huge Learning-focused Approaches} 
& 
& \multirow{3}{*}{\Huge Self-Supervised Learning} 
& \Huge Contrastive Learning             
& TNC \cite{tonekaboni2021unsupervised}, NCL \cite{yeche2021neighborhood}, COMET \cite{wang2023contrast} \\
&
&
& \Huge Predictive and Masked Modeling
& SwitchTab \cite{wu2024switchtab}, MCM \cite{yin2024mcm} \\
&
&
& \Huge Prototype-based Learning
& ProtoMix \cite{xu2024protomix}, SLDG \cite{wu2023iterative} \\
\cmidrule(lr){4-5}
&
& \Huge Clustering-Based
&     
& 
phenotypic clustering \cite{raverdy2024data}, COPD \cite{lee2019dynamic}, \cite{kiyasseh2021crocs}, \cite{seedat2022dataiq}, TACCO \cite{zhang2024tacco}, \cite{carr2021longitudinal}, \cite{landi2020deep}, \cite{qiu2025deep}, \cite{pan2025identification}, ProtoGate \cite{jiang2024protogate}, AC-TPC \cite{lee2020temporal}, \cite{chen2022clustering}, \cite{aguiar2022learning}
\\ 
&
&\Huge Latent Representation Learning
&
& \cite{xue2023assisting}, MixEHR \cite{li2020inferring}, TVAE \cite{xue2023assisting}, ManyDG \cite{yang2023manydg}, TableLM \cite{yang2022numerical}
\\
&
& \Huge Continual Learning
&
& CLOPS \cite{kiyasseh2021clinical}, BrainUICL \cite{zhoubrainuicl}, \cite{armstrong2022continual}, \cite{choudhuri2023continually}
\\
\cmidrule(lr){1-5} 
\multirow{9}{*}{\Huge Multimodal Learning} 
&\multirow{3}{*}{\Huge Alignment in Medical vision-language Models}
&\Huge Global Alignment Strategies.
& 
&
HAIM \cite{lu2024multimodal}, \cite{zhang2022contrastive}, \cite{boecking2022making}, \cite{tiu2022expert}, \cite{huang2023visual}, \cite{zhou2023advancing}, \cite{zhang2023biomedclip}, \cite{lin2023pmc}, \cite{liu2023imitate}, \cite{bannur2023learning}, \cite{cheng2023prior}, \cite{liu2023m}, \cite{sun2024pathasst}, \cite{lu2024visual}, \cite{christensen2024vision}, \cite{lu2024multimodal}, \cite{zhu2024mmedpo}, \cite{chang2025focus}, \cite{lai2023faithful}, \cite{song2024multimodal}, \cite{hemker2024healnet}, \cite{ma2024eye}, \cite{soenksen2022integrated}
\\
\cmidrule(lr){4-5}
&
&\Huge \multirow{3}{*}{Fine-grained Alignment}
&\Huge Extending 2D to 3D Imaging 
& BIUD \cite{cao2024bootstrapping}, CT-CLIP \cite{hamamci2024foundation}, M3D \cite{bai2024m3d}, lin2024ct \cite{lin2024ct}, Merlin \cite{blankemeier2024merlin}, shakya2023benchmarking \cite{shakya2023benchmarking}, BrainMD \cite{wang2024enhancing}
\\
&
&
&\Huge Region-Level Medical LMMs
& Shikra \cite{shikra}, GPT4RoI \cite{gpt4roi}, RegionGPT \cite{regiongpt}, BiRD \cite{bird}, MAIRA-2 \cite{maira}, Med-ST \cite{yang2024unlocking}, fVLM \cite{shui2025large}, MLMMs \cite{li2025aor}, PIBD \cite{zhangprototypical}, \cite{chen2023fine}, UMedLVLM \cite{zhou2025training}, Malenia \cite{jiang2024unleashing}
\\
&
&
&\Huge Fine-Grained Contextual Understanding
& MedKLIP \cite{wu2023medklip}, KAD \cite{zhang2023knowledge}, Imitate \cite{liu2023imitate}, HSCR \cite{jiang2025hscr}, StructuralGLIP \cite{yangprompt}, \cite{liuinterpretable}, TabPedia \cite{zhao2024tabpedia}
\\
\cmidrule(lr){4-5}
&
&\multirow{2}{*}{\Huge \shortstack{Data-Efficient Parallel \\ and Unpaired Alignment}}
&\Huge Parallel Data Collection and Construction
& LVM-Med \cite{mh2023lvm}, HealthGPT \cite{lin2025healthgpt}, Biomedclip \cite{zhang2023biomedclip}, LLaVA-Med \cite{li2023llava}, Med-Flamingo \cite{moor2023med}, OpenFlamingo-9B \cite{awadalla2023openflamingo}
\\
&
&
&\Huge Learning from unpaired Data
& MedCLIP \cite{wang2022medclip}, PTUnifier \cite{chen2023towards}, PairAug \cite{xie2024pairaug}, Med-MLLM \cite{liu2023medical}, M3Care \cite{zhang2022m3care}, Muse \cite{wu2024multimodal}
\\
\cmidrule(lr){2-5}
& \Huge   Temporal Modeling in Medical Vision-Language Model
&
& 
& DDL-CXR \cite{yao2024addressing}, \cite{bannur2023learning}, \cite{liuinterpretable}, Med-ST \cite{yang2024unlocking}, \cite{huo2025timeto}, \cite{zhang2023improving}
\\
\cmidrule(lr){2-5}
&\Huge \multirow{3}{*}{Learning from External Knowledge}
& \Huge Learning from External Databases
& 
& AOR \cite{li2025aor}, \cite{liu2024zero}, StructuralGLIP \cite{yangprompt}
\newline
\textcolor{red}{M3care \cite{zhang2022m3care},
pathchat \cite{lu2024multimodal},
UMed-LVLM \cite{zhou2025training}}
\\
&
& \Huge Internalized Knowledge from LLMs
& 
& EMERGE \cite{zhu2024emerge}, LLM-CXR \cite{lee2024llm-cxr}
\\
&
& \Huge Case-Based Knowledge from Patient Records
& 
& Re3Writer \cite{liu2022retrieve}
\\
\midrule
\multirow{24}{*}{\Huge LLM-Based Modeling and Systems}
&\multirow{9}{*}{\Huge Learning with LLMs}
& \multirow{3}{*}{\Huge Prompt-Based Methods}     
& \Huge Zero/Few-Shot Prompting
&   
DearLLM \cite{xu2025dearllm}, FeatLLM \cite{han2024featllm}, CAAFE \cite{hollmann2023large}, HeartLang \cite{jin2025hegta}, \cite{alsentzer2023zero}, \cite{swaminathan2023natural}
\newline
\textcolor{red}{added \cite{zhang2025exploring,lin2025training,schmidgall2024agentclinic}}
\\
&
&
& \Huge Structured Prompting
& CoK \cite{lichain}, ChAIN-OF-TABLE \cite{wangchain}, Trompt \cite{chen2023trompt}
\newline
\textcolor{red}{Clinical CoT \cite{kwon2024large}}
\\
\cmidrule(lr){3-5} 
& 
&\multirow{3}{*}{\Huge Pretraining and Fine-tuning Methods}      
& 
& 
DiaLLMs \cite{ren2025diallms}
MedFound \cite{liu2025generalist},
Me-LLaMA \cite{xie2025medical},
Med-PaLM2 \cite{singhal2025toward},
Meerkat \cite{kim2025small},
MOTOR \cite{steinbergmotor},
UniTabE \cite{yang2024unitabe},
\cite{abbaspourazad2024wearable},
TP-BERTa \cite{yan2024making},
AMIE \cite{mcduff2025towards},
MERA \cite{ma2025memorize},
\cite{chen2023parameterizing},
GatorTron \cite{yang2022large}
\\
\cmidrule(lr){3-5}
&
&\Huge Retrieval-Augmented Methods
&
& TableRAG \cite{chen2024tablerag},
OpenTable \cite{kongopentab},
Re$^{3}$Writer \cite{liu2022retrieve},
\cite{kresevic2024optimization},
\cite{wang2024prompt},
\cite{borchert2025high},
\cite{ke2025retrieval} 
\newline
\textcolor{red}{added \cite{jeong2024improving},
ClinRaGen \cite{niu2024multimodal},
MedGraphRAG \cite{wu2024medical},
Hykge \cite{jiang2023hykge},
Omni-RAG \cite{chen2025towards},
HSCR \cite{jiang2025hscr}} 
\\
\cmidrule(lr){2-5}
&\multirow{15}{*}{\Huge LLM-Driven Medical Agents}
& \multirow{3}{*}{\Huge{Memory Mechanism}}
& \Huge{Short-term Memory} 
& EHR2Path \cite{pellegrini2025ehrs}, DiagnosisGPT \cite{chen2024cod} 
\\
& 
& 
& \Huge{Long-term Memory} 
& EHRAgent \cite{shi2024ehragent} 
\\
& 
& 
& \Huge{Knowledge Retrieval as Memory}
& ColaCare \cite{wang2025colacare}, MedAgent-Pro \cite{wang2025medagent} 
\\
\cmidrule(lr){3-5}
& 
& \multirow{3}{*}{\Huge{Planning and Reasoning}} 
& \Huge{Static Planning} 
& EHR2Path \cite{pellegrini2025ehrs}, DiagnosisGPT \cite{chen2024cod}, CT-Agent \cite{mao2025ct}, MedRAX \cite{fallahpour2025medrax} 
\\
& 
& 
& \Huge{Dynamic Planning} 
& ColaCare \cite{wang2025colacare}, MedAgent-Pro \cite{wang2025medagent}, MMedAgent \cite{li2024mmedagent}, EHRAgent \cite{shi2024ehragent} 
\\
& 
& 
& \Huge{Iterative Planning} 
& MedAgent-Pro \cite{wang2025medagent}, EHRAgent \cite{shi2024ehragent} 
\\
\cmidrule(lr){3-5}
& 
& \multirow{3}{*}{\Huge{Action Execution}} 
& \Huge{Tool-Based Execution} 
& ColaCare \cite{wang2025colacare}, MedAgent-Pro \cite{wang2025medagent}, MMedAgent \cite{li2024mmedagent}, MedRAX \cite{fallahpour2025medrax} 
\\
& 
& 
& \Huge{Code-Based Execution} 
& EHRAgent \cite{shi2024ehragent} 
\\
& 
& 
& \Huge{Environment-Embedded Execution} 
& MedAgentBench \cite{jiang2025medagentbench}, \textcolor{red}{AgentClinic \cite{schmidgall2024agentclinic}}
\\
\cmidrule(lr){3-5}
& 
& \multirow{3}{*}{\Huge{Self-Improvement}} 
& \Huge{Memory-Based Self-Improvement} 
& EHRAgent \cite{shi2024ehragent} 
\\
& 
& 
& \Huge{Interactive Refinement} 
& MedAgent-Pro \cite{wang2025medagent}, DiagnosisGPT \cite{chen2024cod} 
\\
& 
& 
& \Huge{Tool Adaptation and Extension} 
& MMedAgent \cite{li2024mmedagent}, ColaCare \cite{wang2025colacare}, AgentClinic \cite{schmidgall2024agentclinic} 
\\
\cmidrule(lr){3-5}
& 
& \multirow{3}{*}{\Huge{Multi-Agent Collaborations}} 
& \Huge{Centralized Control} 
& ColaCare \cite{wang2025colacare}, MedAgent-Pro \cite{wang2025medagent} 
\\
& 
& 
& \Huge{Decentralized Collaboration} 
&  
\\
& 
& 
& \Huge{Hybrid Architectures} 
& MMedAgent \cite{li2024mmedagent} 
\\
\midrule
\multirow{17}{*}{\Huge Clinical Applications} 
& \multirow{7}{*}{\Huge Clinical Document Understanding} 
& \Huge{Clinical Document Summarization} 
&
& \cite{yackel2010unintended}, \cite{bowman2013impact}, \cite{ghosh2024sights}, \cite{wang2020cord}, \cite{deyoung2021ms2}, \cite{wallace2021generating}, \cite{joshi2020dr}, \cite{poornash2023aptsumm}, BART \cite{lewis2019bart}, PEGASUS \cite{zhang2020pegasus}, ProphetNet \cite{qi2020prophetnet}, \cite{goff2018automated}, \cite{rajaganapathy2025synoptic}, \cite{van2024adapted}, \cite{asgari2025framework} \\
& & \Huge{Clinical Note Generation}  
&
& Flamingo-CXR \cite{tanno2025collaboration} \\
& & \Huge{Concept Extraction}  
&
& MetaMap \cite{aronson2010overview}, cTAKES \cite{savova2010mayo}, MedLEE \cite{friedman1994general}, CLAMP \cite{soysal2018clamp}, ScispaCy \cite{neumann2019scispacy}, medspaCy \cite{eyre2022launching}, EHRKit \cite{li2022ehrkit}, UmlsBERT \cite{michalopoulos2020umlsbert}, CancerBERT \cite{zhou2022cancerbert}, CLEAR \cite{lopez2025clinical}, \cite{keloth2025social}, \cite{hein2025iterative} \\
& & \Huge{Image-Text Retrieval}  
&
& MONET \cite{kim2024transparent}, \cite{huang2024critical} \\
& & \Huge{Clinical Coding Automation}  
&
& \cite{li2020icd}, \cite{xie2019ehr}, \cite{cao2020hypercore}, \cite{yuan2022code}, \cite{dong2021rare}, \cite{gao2022classifying}, BERT \cite{devlin2019bert}, \cite{pascual2021towards}, \cite{huang2022plm}, Longformer \cite{beltagy2020longformer}, \cite{yang2022knowledge}, \cite{lu2023towards}, CliniCoCo \cite{gao2024optimising} \\
& & \Huge{Quantitative Clinical Calculations}  
&
& \cite{goodell2025large} \\
& & \Huge{General Clinical AI Capabilities}  
&
& \cite{lu2024visual}, \cite{chen2024towards}, \cite{christensen2024vision}, \cite{yu2024heterogeneity}, \cite{groh2024deep} \\
\cmidrule(lr){3-5}
& \multirow{7}{*}{\Huge Clinical Reasoning and Decision Support} 
& \multirow{2}{*}{\Huge{Diagnosis Prediction}}
& --
& \cite{ng2023prospective}, \cite{cao2023large}, \cite{thieme2023deep}, \cite{yao2025multimodal} \\
&
& 
& \Huge{Differential Diagnosis}
& \cite{xue2024ai}, \cite{mao2025phenotype}, \cite{fast2024autonomous}, \cite{chen2025enhancing}
\\
\cmidrule(lr){4-5}
& & \multirow{3}{*}{\Huge{Prognostic Forecasting}}
&\Huge{Risk Prediction}
&RETAIN \cite{choi2016retain}, T-LSTM \cite{baytas2017patient}, \cite{dai2024deep}, \cite{garriga2022machine}, \cite{luo2020hitanet}, MCA-RNN \cite{lee2018diagnosis}, SAnD \cite{song2018attend}, ConCare \cite{ma2020concare}, GRAM \cite{choi2017gram}, KAME \cite{ma2018kame}, HAP \cite{zhang2020hierarchical}, GRASP \cite{zhang2021grasp}, \cite{chen2024predictive}\\
& &
& \Huge{Readmission Prediction}
& AutoDP \cite{cui2024automated}, SMART \cite{yu2024smart}, \cite{wu2024instruction}\\
& &
& \Huge{Length of Stay \& Mortality Prediction}
& DeepMPM \cite{yang2022deepmpm}, TECO \cite{rong2025deep}, ChronoFormer \cite{zhang2025chronoformer}\\
\cmidrule(lr){4-5}
& & \multirow{2}{*}{\Huge{Cohort Discovery}  }
& \Huge{Cohort Discovery}
& COMET \cite{mataraso2025machine}\\
& &
& \Huge{Subtyping/Endotyping}
& \cite{raverdy2024data}, Virchow \cite{vorontsov2024foundation}, AFISP \cite{subbaswamy2024data}\\
\cmidrule(lr){3-5}
& \multirow{3}{*}{\Huge Clinical Operations Support} 
& \Huge{Triage and Prioritization}  
&
& AISmartDensity \cite{salim2024ai}, \cite{chang2025continuous}, \cite{gaber2025evaluating} \\
& & \Huge{Referral recommendation}  
&
& \cite{habicht2024closing}, DeepDR-LLM \cite{li2024integrated}, SSPEC \cite{wan2024outpatient}, \cite{groh2024deep} \\
& & \Huge{Patient–Trial Matching}  
&
& PRISM \cite{gupta2024prism}, \cite{alkhoury2025enhancing} 
\\
\midrule
\multirow{3}{*}{\Huge Evaluation Benchmark} 
& \Huge{Benchmark}
& 
& 
& \cite{goh2025gpt}, \cite{sharma2023human}
\\
\cmidrule(lr){3-5}
& \multirow{2}{*}{\Huge{Datasets}}
& \Huge{Tabular EHR}
&
& \cite{johnson2016mimic}, \cite{johnson2023mimic}, \cite{pollardEICUCollaborativeResearch2018}, \cite{wornowEhrshotEhrBenchmark2023}, \cite{al-arsNLICESyntheticMedical2023}
\\
& 
& \Huge{Text-based EHR}
&
& \cite{flemingMedalignCliniciangeneratedDataset2024}, \cite{jinWhatDiseaseDoes2020}, \cite{chenBenchmarkingLargeLanguage2025}, \cite{vilaresHEADQAHealthcareDataset2019}, \cite{zhouExplainableDifferentialDiagnosis2025}, \cite{fansitchangoDdxplusNewDataset2022}, \cite{abacha2025medecbenchmarkmedicalerror}, \cite{khandekar2024medcalc}, \cite{jin2019pubmedqa}, \cite{kimMdagentsAdaptiveCollaboration2024}, \cite{panditMedHalluComprehensiveBenchmark2025}, \cite{mtsamples}, \cite{xuOverviewFirstShared2024}, \cite{johnson2023mimicivnote}, \cite{yim2023aci}, \cite{abachaBridgingGapConsumers2019}, \cite{medlineplus}
\\
\bottomrule
\end{tabular}
}
\caption{Coarse-to-Fine Taxonomy of Deep Learning Design Elements for EHR}
\label{tab:design_taxonomy}
\end{table*}

%% file: Figures/section3/figure_section3.tex
\begin{figure}[htbp]
  \centering
  \includegraphics[width=\linewidth]{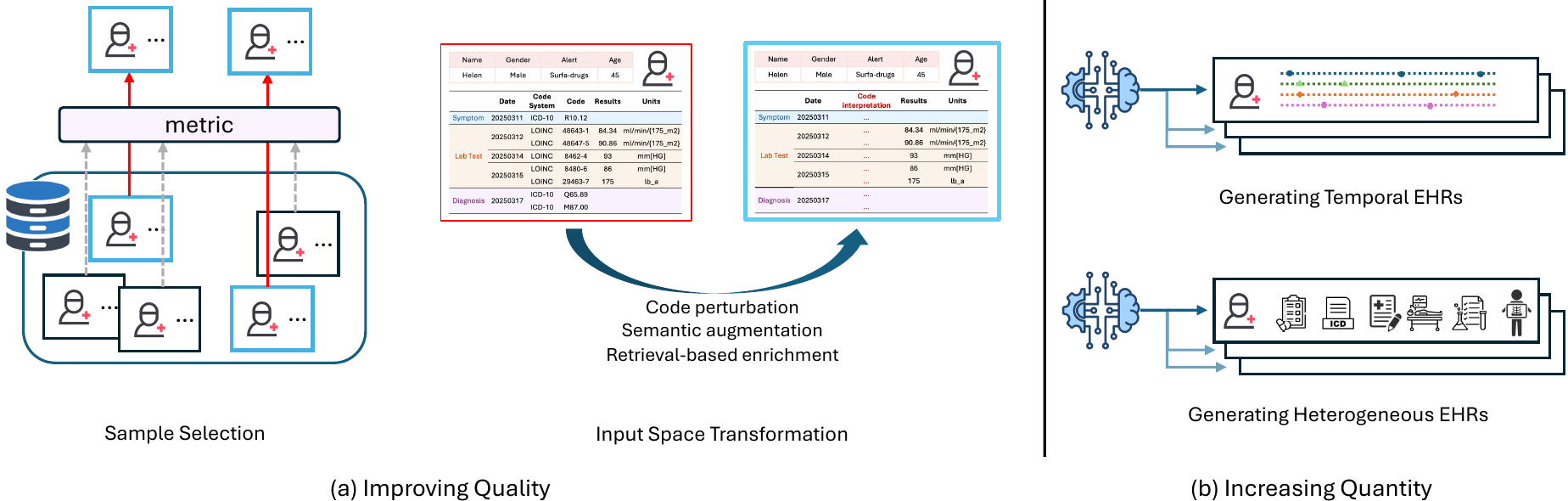}
  \caption{An illustration of data-centric approaches.}
  \label{fig:data-centric}
\end{figure}

%% file: Sections/Methodology/Neural_Architectural_Approaches.tex
\section{Neural Modeling Strategies}
\label{neural_model}
Neural modeling strategies lie at the core of deep learning approaches for EHRs, focusing on designing architectures that can effectively handle heterogeneity, temporal irregularity, hierarchical structure, and domain specificity inherent in clinical data. We categorize these strategies into four main branches: (1) \textbf{feature-aware modules}, (2) \textbf{structure-aware architecture design}, (3) \textbf{temporal dependency modeling}.
\subsection{Feature-aware Design}
To accommodate the diverse data types in EHRs, ranging from binary indicators to high dimensional continuous measurements, feature aware modules are designed to preprocess and encode inputs in a manner that enhances both generalization and interpretability.
\subsubsection{Discretization and Binning-Based Methods}

\input{Figures/section4/figure_section4_sep1}
The heterogeneous nature of EHRs presents several challenges to deep learning models. First, EHRs contain diverse feature types, including boolean, categorical, and numerical variables, each with distinct underlying distributions.
Second, the dependencies among these features are often complex and not known a priori, making it difficult for models to learn accurate representations without strong inductive biases.
Recently, one emerging research direction involves discretizing numerical features into categorical representations, aiming to simplify modeling and enhance the capacity of neural networks to capture non-linear patterns.
For example,
Lee et al. \cite{lee2024binning} employed discretization as a self-supervised pretext task, leveraging data-driven binning to enhance neural representation learning. 
Ren et al. \cite{ren2024tablog} proposed TabLog, utilizing logic-based discretization for robust test-time adaptation, effectively addressing distribution shifts common in EHR scenarios.
Karimireddy et al. \cite{karimireddy2020byzantine} developed a bucketing approach to manage heterogeneous datasets, improving robustness against corrupted data by grouping similar data points.
Gorishniy et al. \cite{gorishniy2022embeddings} examined numerical embeddings through discretization, demonstrating enhanced predictive performance via categorical embeddings of continuous variables.
Zantedeschi et al. \cite{zantedeschi2021learning} integrated discretization into differentiable binary decision trees, facilitating gradient-based optimization and improving both interpretability and predictive accuracy.

\subsubsection{Kernel-Based Methods}
Kernel-based methods provide a powerful and flexible framework for modeling complex, non-linear relationships in EHRs.
By implicitly mapping input features into high-dimensional spaces, kernel techniques enable models to learn rich feature interactions that are not easily captured by standard linear architectures. 
Good et al. \cite{good2023feature}  introduced a method for learning interpretable and performant decision trees by integrating kernel-inspired feature learning. Their approach constructs smooth, differentiable transformations of the input space, enabling decision trees to approximate non-linear boundaries while preserving interpretability.

\subsection{Structure-aware Design}

\input{Figures/section4/figure_section4_sep2}
Electronic health records exhibit inherent structural properties, such as hierarchical coding systems, interdependent features, and visit-based temporal organization. Structure-aware architecture design leverages these properties to guide model inductive biases, enhance data efficiency, and support clinically meaningful representations.
\subsubsection{Tree-based}
A tree is a hierarchical, acyclic data structure that organizes information through parent–child relationships, with a single root and uniquely connected nodes. In EHR data, tree structures arise naturally and offer a powerful inductive bias to incorporate clinical knowledge, reduce data sparsity, and improve interpretability. Tree-based modeling in EHRs is primarily applied in three scenarios:
(1) \textbf{Tree-Guided Feature Selection},
(2) \textbf{Tree-Guided Clinical Coding},
and (3) \textbf{Tree-Based Event Sequence Modeling}.

\underline{\textit{Tree-Guided Feature Selection.}}
Due to the high dimensionality and heterogeneity of EHR data, tree-based models such as decision trees, random forests, and gradient-boosted trees are widely used for feature selection \cite{murris2024tree,chen2024tree}. These models naturally handle nonlinear interactions and provide interpretable decision rules \cite{chen2016xgboost}. Furthermore, domain-specific hierarchies can be incorporated as tree-structured regularizers to enforce sparsity and improve semantic coherence \cite{good2023feature}.
\underline{\textit{Tree-Guided Clinical Coding}}. Medical ontologies such as ICD (for diagnoses), CPT (for procedures), and ATC (for medications) are inherently tree-structured, encoding semantic hierarchies where internal nodes represent broad categories and leaves denote specific clinical concepts. 
Tree-Guided Clinical Coding has been used to extract structured codes from unstructured clinical text by aligning token representations with the hierarchical structure of the target codes \cite{liu2022treeman,lu2023towards}.
\underline{\textit{Tree-Based Event Sequence Modeling.}}
In modeling patient trajectories, tree structures can represent latent decision pathways or logic programs that explain the temporal progression of clinical events. Existing works \cite{pokharel2020representing,song2024latent} model each event as a node and construct temporal trees to capture causal or logical dependencies, enabling interpretable modeling of care trajectories.

\subsubsection{Graph-based}
EHRs contain complex, multi-scale clinical information encompassing diagnoses, procedures, medications, and temporally ordered visit sequences. Graph-based modeling provides a flexible and structured framework to capture the rich relational and hierarchical dependencies \cite{li2022graph}.
This modeling paradigm can be applied at multiple levels of abstraction, including (1) \textbf{code level}, (2) \textbf{visit level}, (3) \textbf{patient level}, and (4) \textbf{hybrid level}.

\underline{\textit{Code-Level Graph Modeling.}}
At the most granular level, graph-based models aim to learn structured representations of individual medical codes. In these models, nodes represent clinical concepts (e.g., ICD, CPT, ATC codes), and edges encode semantic relationships derived from medical ontologies or empirical co-occurrence statistics. For example, GRAM \cite{choi2017gram} leverages hierarchical attention over medical ontologies, while G-BERT \cite{shang2019pre} applies GNNs to embed medical codes and integrates them into language models for downstream prediction tasks.
\underline{\textit{Visit-Level Graph Modeling.}}
Clinical visits often involve sets of co-occurring medical codes that do not follow a strict temporal order. Visit-level graph models construct graphs in which nodes correspond to codes within a visit, and edges capture statistical co-occurrence, functional relationships, or learned dependencies. The Graph Convolutional Transformer (GCT) \cite{choi2020learning} represents each visit as a fully connected graph over its codes, enabling the model to capture structured intra-visit interactions that are overlooked by purely sequential approaches \cite{zhu2021variationally}.
\underline{\textit{Patient-Level Graph Modeling.}}
At a higher level, graphs are used to represent temporal dynamics and population-wide relationships. Here, nodes may represent patients, visits, or aggregated temporal features \cite{lu2021collaborative}, with edges encoding temporal transitions, clinical similarity, or interactions among entities. ME2Vec \cite{wu2021leveraging} constructs a heterogeneous graph linking patients, services, and providers.
\cite{park2022graph} proposes a multimodal approach with a GNN to extract the graphical structure of EHR and a language model to handle clinical notes.
\cite{lu2022context} constructs a global disease co-occurrence graph with multiple node properties for disease combinations. We design dynamic subgraphs for each patient's visit to leverage global and local contexts.
\cite{li2025time} employs a co-guided graph to connect various patient sequences.
HI-DR \cite{kim2025hi} and funGCN \cite{boschi2024functional} leverages GNNs to propagate patient status.
\underline{\textit{Hybrid-Level Graph Modeling.}}
Hybrid models aim to capture dependencies across multiple levels of the EHR structure. These approaches integrate code-level ontologies, visit-level graphs, and patient-level temporal sequences into a unified framework. 
For instance, \cite{lee2020harmonized}  apply GNNs to visit graphs and use their aggregated outputs as inputs to temporal models.
\cite{poulain2024graph} dynamically incorporates external knowledge graphs into longitudinal patient trajectories. Such hybrid architectures enable rich multi-level reasoning and often lead to improved accuracy, generalizability, and interpretability in complex clinical prediction tasks.

\subsubsection{Rule-based Methods}
EHRs capture clinical processes that are inherently structured and governed by domain knowledge, including diagnostic criteria, treatment protocols, temporal dependencies, and physiological constraints. Rule-based methods aim to incorporate this knowledge into learning and inference by encoding symbolic rules, logical constraints, or procedural logic. By aligning model behavior with established clinical reasoning, these approaches enhance trustworthiness, improve generalization, and promote consistency \cite{lowe2022logical}.
Recent advances demonstrate the utility of integrating rule-based reasoning with tabular-based EHRs. For instance, SQRL \cite{naik2023machine} combines logic-based constraints with statistical inference to induce rules from data in an unsupervised manner. Tablog \cite{ren2024tablog} employs a logic neural network to learn fuzzy-logic rules over tabular data. ConSequence \cite{theodorou2024consequence} enhances EHR generation by incorporating rule-based priors, improving realism and domain alignment.

\subsubsection{Neural Additive Models (NAMs)}
EHRs consist of structured, heterogeneous clinical variables such as demographics, diagnoses, lab results, and medications. These features are often interpreted independently in clinical practice and contribute additively to risk assessments and treatment decisions.
NAMs \cite{agarwal2021neural} offer a suitable modeling framework by combining the interpretability of additive models with the flexibility of neural networks \cite{cheng2024arithmetic}.
Formally, NAMs predict outcomes via an additive structure:
\begin{equation}
    \hat{y} = \beta_0 + \sum_{j=1}^p f_j(x_j)
\end{equation}
where each $f_j$ is a neural network that models the contribution of feature $x_j$, and $p$ is the number of features.
This form enables learning complex, non-linear effects while preserving feature-level attribution.
For example, SIAN \cite{enouen2022sparse} introduces a feature selection mechanism to identify and incorporate a sparse set of statistically relevant feature combinations. This allows the model to capture complex dependencies while preserving an additive and interpretable structure \cite{ibrahim2023grand,lyu2023towards}. 

\subsubsection{Hierarchical and Structured Transformers
}
Transformer-based models \cite{vaswani2017attention} have been widely adapted to EHR data due to their capacity to model both temporal structure, such as sequences of clinical visits, and hierarchical composition, where each visit contains multiple medical codes. These models typically treat either visits or medical codes as input tokens, enabling them to capture different levels of clinical relationships \cite{wang2024transformer}.

One line of work models visits as tokens, where each visit is represented by an embedding derived from its constituent medical codes. HiTANet \cite{luo2020hitanet} uses a fully connected layer to generate visit-level embeddings and applies time-aware attention to capture both local and global temporal dependencies. RAPT \cite{ren2021rapt} follows a similar approach for visit representation but focuses on improving patient modeling through task-specific pretraining objectives.
Another line of work treats medical codes as input tokens, focusing on modeling fine-grained clinical semantics within and across visits. Models such as TransformEHR \cite{yang2023transformehr}, BEHRT \cite{li2020behrt}, MedBERT \cite{rasmy2021med}, RareBERT \cite{prakash2021rarebert}, CEHR-BERT \cite{poulain2022few}, and BEHRT-CVD \cite{poulain2021transformer} adopt transformer architectures to capture contextual relationships among codes. These models enrich input representations with temporal, demographic, and domain-specific embeddings to enhance both predictive accuracy and interpretability.
Recent studies focus on capturing both code-level and visit-level semantics. DL-BERT \cite{chen2022dl} encodes intra-visit and inter-visit information by stacking two BERT-based modules. GT-BEHRT \cite{poulain2024graph} integrates graph-based visit embeddings with a sequential BERT model, allowing the architecture to represent both structured relationships among medical concepts and temporal dependencies across visits.

\subsection{Temporal Dependency Modeling}

\input{Figures/section4/figure_section4_sep3}
\subsubsection{Irregular/Asynchronous Sampling}
Irregular sampling is a core challenge in modeling EHR time series due to heterogeneous visit intervals, missing values, and asynchronous measurements across clinical variables. Existing methods can be broadly categorized into two paradigms:
\textbf{(1) Implicit Temporal Handling.}
These approaches incorporate temporal dynamics directly into model architectures without explicitly aligning timestamps \cite{chung2020deep}. GRU-D \cite{che2018recurrent} augments recurrent units with decay terms and missingness masks. Neural ODEs \cite{chen2018neural} model continuous-time dynamics via differential equations, enabling flexible handling of arbitrary time gaps. SeFT \cite{horn2020set} and ContiFormer \cite{chen2023contiformer} extend Transformers with time-aware attention mechanisms. DNA-T \cite{huang2024dna} further introduces deformable attention to dynamically adjust receptive fields for local patterns. Graph-based methods \cite{zhang2021graph, yin2024sepsiscalc,chen2024predictive} construct dynamic temporal graphs to model evolving interactions among clinical events.
Warpformer \cite{zhang2023warpformer} integrates both perspectives by learning multi-scale temporal representations through adaptive warping, effectively balancing coarse- and fine-grained views.
Unlike previous
Recently works \cite{li2023time,chen2025integrating}  transforms the signals into RGB images and utilizes a pre-trained Swin Transformer for further
classification and regression.
\textbf{(2) Explicit Temporal Alignment.}
These methods convert irregular sequences into regular representations \cite{katsuki2022cumulative,li2021imputation}. One strategy interpolates observations to fixed reference points \cite{shukla2019interpolation,shukla2021multi,zhang2023improving,zhong2024hybrid}, while another unfolds sequences into $(\text{value}, \text{type}, \text{time})$ tuples to preserve fine-grained temporal granularity \cite{tipirneni2022self}. 

\subsubsection{Multi-Timescale Dynamics}
Clinical data often exhibit dynamics that unfold over multiple temporal resolutions \cite{nguyen2021clinical,ma2020adacare}. For example, acute events such as sudden changes in vital signs occur over minutes or hours, while chronic conditions and long-term disease progression span weeks to years. Capturing both short-term fluctuations and long-term trends is thus crucial for robust EHR modeling \cite{shi2023modelling}.
Early efforts to handle multiscale information used hierarchical RNNs or stage-aware models \cite{li2020deepalerts}. For instance, StageNet \cite{gao2020stagenet} decomposes patient records into clinical stages and models transitions with separate modules, explicitly capturing stage-level temporal structure. 
Autoformer \cite{wu2021autoformer} and FedFormer \cite{zhou2022fedformer} utilize decomposition and frequency-domain filtering to capture seasonal and trend components at various scales. Although these were originally developed for regular time series, their principles have inspired adaptations to EHRs settings. 
DeepAlerts \cite{li2020deepalerts} treating each prediction horizon as a task at a different time scale and encourage the model to capture inter-horizon temporal correlations.
AdaCare \cite{ma2020adacare} capture the long and short-term variations of biomarkers as clinical features to depict the health status in multiple time scales.

\subsubsection{Conditional Clinical Sequences}
Medical events in EHRs unfold in a conditional manner, where future decisions depend on previously observed symptoms, diagnoses, and treatments \cite{placido2023deep}. This dynamic reflects the step-wise reasoning process inherent in clinical workflows \cite{chen2021novel}. To model such dependencies, recent works have formulated clinical prediction as a conditional sequence generation task, aligning with how clinicians synthesize context to make informed decisions.
DeepAlerts \cite{li2020deepalerts} generates multi-horizon deterioration alerts conditioned on patient history and time-to-event proximity. It treats each prediction horizon as a related subtask and leverages task-related priors to optimize performance jointly.
Similarly, Diaformer \cite{chen2022diaformer} reformulates diagnosis as a symptom sequence generation task, modeling the diagnostic process as conditional sequence prediction rather than static classification. It introduces symptom-aware attention and orderless training to accommodate the flexible ordering and dynamic reasoning inherent in clinical decision-making.
MICRON \cite {yang2021change} captures context dependency by recurrently updating a latent medication state with new clinical inputs, enabling efficient, temporally aware prediction of medication changes.
ConCare \cite{ma2020concare} focuses on learning personalized, context-dependent representations by conditioning predictions on both static patient features and dynamically evolving medical histories. 
DiaLLMs \cite{ren2025diallms} employs a reinforcement learning framework for evidence acquisition and automated diagnosis recommendation.
These models emphasize capturing conditional dependencies in clinical sequences by aligning predictions with the temporal and contextual logic underlying clinical decision-making, treatment planning, and risk stratification.

%% file: Figures/section4/figure_section4_sep1.tex
\begin{wrapfigure}{r}{0.45\textwidth}  
  \centering
  \vspace{-11pt} \includegraphics[width=0.45\textwidth]{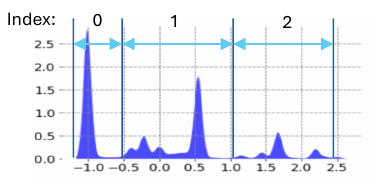}
  \caption{An illustration of binning methods.}
  \vspace{-12pt} 
  \label{binning}
\end{wrapfigure}

%% file: Figures/section4/figure_section4_sep2.tex
\begin{figure}[htbp]
  \centering
  \includegraphics[width=1\textwidth]{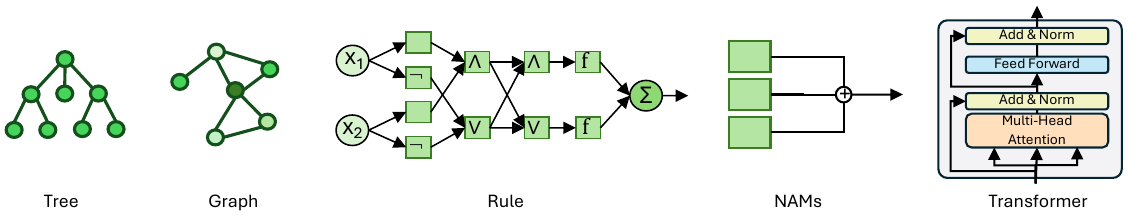}
  \caption{An illustration of structure-aware designs.}
  \label{structure aware}
\end{figure}

%% file: Figures/section4/figure_section4_sep3.tex
\begin{figure}[htbp]
  \centering
  \includegraphics[width=1\textwidth]{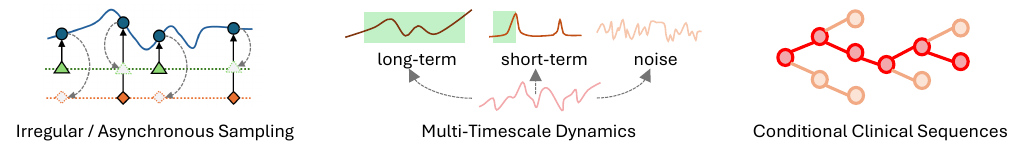}
  \caption{An illustration of temporal dependency modeling strategies.}
  \label{temporal modeling}
\end{figure}

%% file: Sections/Methodology/Learning_focused.tex
\section{Learning-Focused Approaches}
\label{objective}
\subsection{Self-Supervised Learning for EHR Data}
Self-supervised learning (SSL) has emerged as a promising paradigm for modeling EHRs by exploiting intrinsic data patterns without requiring manual annotations. By leveraging the temporal structure, clinical codes, and co-occurrence patterns inherent in EHRs, SSL enables the learning of rich patient representations that generalize across downstream clinical tasks such as risk prediction, diagnosis classification, and patient trajectory modeling. This approach facilitates scalable and data-efficient learning, addressing the challenge of limited labeled data in healthcare.

\subsubsection{Contrastive Learning}
Contrastive learning encourages representation alignment between augmented views of the same instance while pushing apart those from different instances, typically using an instance discrimination objective. In the EHR domain, TNC \cite{tonekaboni2021unsupervised} and NCL \cite{yeche2021neighborhood} exploit local temporal structures by contrasting neighboring and non-neighboring time points to capture trial-level consistency. COMET \cite{wang2023contrast} extends this approach through hierarchical contrastive learning at multiple levels—observation, sample, trial, and patient—enabling the extraction of semantically rich representations from medical time series without requiring expert-labeled data.
\subsubsection{Predictive and Masked Modeling}
Masked prediction and reconstruction methods learn EHR representations by recovering masked or corrupted input features, allowing models to capture intrinsic dependencies among clinical variables. Let \( x \in \mathbb{R}^d \) denote an EHR sample and \( M(\cdot) \) be a masking operator applied to selected input dimensions. The training objective minimizes the reconstruction error:
\begin{equation}
\mathcal{L}_{\text{recon}} = \mathbb{E}_{x \sim \mathcal{D}} \left[ \| f_{\theta}(M(x)) - x \|^2 \right],
\end{equation}
where \( f_{\theta} \) is an encoder–decoder architecture parameterized by \( \theta \). This encourages the model to encode latent structures that support robust imputation and generalization.
Recent methods extend this paradigm with structured masking and auxiliary regularization.  SwitchTab~\cite{wu2024switchtab} encodes two randomly corrupted samples $x_1$ and $x_2$ into feature vectors, which are decomposed into mutual ($m_1, m_2$) and salient ($s_2, s_2$) components using two separate projectors. A decoder $d$ then reconstructs both the original and "switched" inputs, encouraging the model to learn disentangled representations. The reconstruction loss is defined as:
$\mathcal{L}_{\text{rec}} = \sum_{i=1}^{2} \left( \| d(s_i \oplus m_i) - x_i \|_2^2 + \| d(s_i \oplus m_{j \neq i}) - x_i \|_2^2 \right).
$
MCM \cite{yin2024mcm} extends masked modeling to anomaly detection by learning intrinsic feature correlations in tabular data, where deviations from these correlations indicate anomalies.

\subsubsection{Prototype-based Learning}
In EHR representation learning, prototype-based self-supervised methods improve generalizability and interpretability by encouraging patient representations to align with latent clinical prototypes. Unlike contrastive learning that emphasizes instance-level discrimination, these methods cluster semantically similar patient records around shared prototype vectors, capturing underlying phenotypic patterns. 
ProtoMix~\cite{xu2024protomix} mitigates the limitations of generative augmentation in low-resource EHR scenarios by employing prototype-guided mixup. It synthesizes samples anchored to clinical subtype centroids, adaptively modulating mixing coefficients to enhance distributional coverage and preserve phenotype structure.
SLDG~\cite{wu2023iterative} introduces a self-supervised domain generalization framework that infers latent patient subgroups and trains subgroup-specific predictors, enabling personalized modeling without explicit domain labels and improving robustness to domain shifts in clinical data.

\subsection{Clustering-based Methods}
Many medical conditions exhibit substantial inter-patient heterogeneity in pathology, disease progression, treatment response, and outcomes \cite{raverdy2024data}. 
An important goal is to identify phenotypically separable clusters with distinct phenotypic profiles (which we denote as phenotypic clustering hereafter) to advance our understanding of disease mechanisms, facilitate phenotype discovery, support personalized prognosis, and inform targeted treatment strategies. 
For instance, in chronic conditions such as Chronic Obstructive Pulmonary Disease (COPD) \cite{lee2019dynamic}, stratifying patients based on exacerbation risk, physiological trajectories, or treatment response has led to improved clinical outcomes and more effective care delivery.
However, clustering EHR data is inherently challenging due to its heterogeneity, temporal irregularity, and complex feature dependencies.

Existing clustering methods leverage diverse patient-level criteria to inform clusters, including static attributes (e.g., disease class, sex, age) \cite{kiyasseh2021crocs}, predictive uncertainty \cite{seedat2022dataiq},  correlation between clinical concepts and patient visits \cite{zhang2024tacco}, and longitudinal medical history \cite{carr2021longitudinal, landi2020deep}, diagnosis trajectories \cite{qiu2025deep}, subphenotypes \cite{pan2025identification} or unsupervised clustering \cite{jiang2024protogate} in a data-driven way.
Another line of work incorporates patient temporal trajectory information into clustering. For example, AC-TPC \cite{lee2020temporal} and \cite{chen2022clustering} stratify patients by learning representations that capture future outcome distributions through outcome-guided loss functions. In contrast, Aguiar et al. \cite{aguiar2022learning} introduce a feature-time attention mechanism to identify phenotype-relevant features across both temporal and feature dimensions.

\subsection{Latent Representation Learning}
Latent representation learning is essential for modeling electronic health records (EHRs), which are inherently sparse, high-dimensional, and heterogeneous. By transforming raw clinical data into compact, informative embeddings, latent variable models uncover hidden structures that facilitate downstream tasks such as phenotype discovery, diagnosis prediction, and treatment effect estimation. 

One line of work learns \underline{\textit{discriminative and disentangled latent representations}} \cite{xue2023assisting}.
For instance, MixEHR~\cite{li2020inferring} represents patients as mixtures of meta-phenotypes, revealing interpretable disease patterns. 
TVAE~\cite{xue2023assisting} learns disentangled latent variables to jointly model treatment assignments and outcomes, addressing selection bias in causal inference. ManyDG~\cite{yang2023manydg} improves generalization to unseen patients by treating each patient as a domain and removing spurious correlations through latent-space orthogonalization. 
Another line of work focuses on \underline{\textit{learning contextual representations}}, such as TableLM \cite{yang2022numerical}, which represents tabular data as sentence-like structures for context understanding.



\subsection{Continual Learning}
EHR data are continuously collected across multiple clinical sites and heterogeneous sensors, resulting in temporal and cross-institutional variability. This induces distribution shifts, compounded by the fact that patient status evolves over time. Continual learning frameworks address these challenges by enabling models to incrementally adapt to new data while retaining previously acquired knowledge, thereby enhancing generalization and robustness in dynamic clinical environments.
Existing works can be categorized into three directions: (1) replay-based method. 
For example, CLOPS \cite{kiyasseh2021clinical}, BrainUICL \cite{zhoubrainuicl} and \cite{armstrong2022continual} employs a replay buffer to store patient historical information. 
(2) Regularization-based methods penalize changes in model parameters to mitigate catastrophic forgetting. For example, \cite{choudhuri2023continually} proposes a temporal consistency loss to learn a generalizable patient embedding.
(3) parameter isolation methods allocate distinct subsets of model parameters to different tasks or domains, reducing interference across learning stages. These approaches often rely on dynamic gating or mask-based mechanisms to isolate knowledge associated with different environments or patient populations. While widely explored in computer vision and NLP, relatively few works have applied parameter isolation techniques to EHR data, presenting an open opportunity for future research in this area.

\section{Multi-Modality in Clinical AI}
\label{sec::multi-modality}
Modern clinical decision-making increasingly relies on the integration of heterogeneous data sources, such as EHRs and medical imaging. EHRs provide structured and temporal information about patient history, laboratory results, medications, and diagnoses, while imaging modalities offer rich anatomical and physiological insights. Combining these complementary sources enables a more comprehensive understanding of patient status and supports more accurate and personalized clinical decisions. Among imaging modalities, computed tomography (CT), magnetic resonance imaging (MRI), digital radiography (DR), and ultrasound (US) are widely used, each with distinct properties in terms of acquisition method, diagnostic value, and applicability. Table 3 summarizes the key characteristics of these modalities to inform the design of multi-modal AI systems capable of holistic patient modeling.
\input{Tables/Multimodality}
\subsection{Alignment in Medical vision-language Pretraining}
Integrating heterogeneous modalities, including medical images, clinical text, structured EHRs, and sensor data, is crucial for building robust and generalizable medical AI systems. 
However, modality-specific semantics, semantic gaps and data incompleteness present significant challenges for effective multimodal alignment.
Medical vision-language pretraining (Med-VLP) methods have primarily focused on 2D imaging, tailored for specific domains such as pathology \cite{lu2024multimodal} and radiology. These models typically learn transferable representations by globally aligning images with associated text \cite{zhang2022contrastive,boecking2022making,tiu2022expert,huang2023visual,zhou2023advancing,zhang2023biomedclip,lin2023pmc,liu2023imitate,bannur2023learning,cheng2023prior,liu2023m,sun2024pathasst,lu2024visual,christensen2024vision,lu2024multimodal,zhu2024mmedpo,chang2025focus,lai2023faithful},
RNA sequence \cite{song2024multimodal,hemker2024healnet}, Radiologist diagnosis response\cite{ma2024eye}.
HAIM \cite{soenksen2022integrated} generates unified patient embeddings by independently encoding tabular, time-series, text, and image data using lightweight transformations and pre-trained models, enabling multimodal predictive analytics with models like XGBoost.

\subsubsection{Fine-Grained Alignment Strategies}
While global alignment captures high-level semantic associations, it fails to model fine-grained correspondences between specific image regions and localized textual descriptions \cite{muller2022joint}, limiting the model’s ability to detect clinically significant spatial features.
To improve fine-grained alignment, recent work advances along two dimensions. 

From \textbf{image-level} perspective, existing work in medical vision-language pretraining (Med-VLP) can be broadly categorized into two major directions:
(1) \textbf{Extending 2D to 3D Imaging}, which enhances anatomical representation and diagnostic fidelity by leveraging volumetric medical data; and
(2) \textbf{Region-Level Medical LMMs}, which focus on extracting and localizing fine-grained visual features to improve spatial grounding and lesion understanding.
\underline{\textit{Extending 2D to 3D Imaging}}.
Recent efforts extend Med-VLP beyond 2D modalities to high-resolution 3D computed tomography (CT), enabling richer anatomical coverage and more detailed diagnostic support \cite{hamamci2024foundation,bai2024m3d,lin2024ct,blankemeier2024merlin,shakya2023benchmarking}. For example, BIUD \cite{cao2024bootstrapping} and CT-CLIP \cite{hamamci2024foundation} align volumetric chest CT data with radiology reports, while Merlin \cite{blankemeier2024merlin} targets abdominal CT and integrates structured EHR data as additional supervision.
BrainMD \cite{wang2024enhancing} selects representative 2D slices from 3D medical image to improve image quality.
\underline{\textit{Region-Level Medical LMMs}}.
To further enhance granularity, recent studies in general-domain LMMs incorporate region-level visual representations. Shikra \cite{shikra} quantizes bounding boxes into coordinate embeddings, while GPT4RoI \cite{gpt4roi} and RegionGPT \cite{regiongpt} embed region features directly into the token sequence, enabling multimodal models to understand localized visual prompts.
In the medical domain, region-level modeling remains underexplored. BiRD \cite{bird} equips medical LMMs with grounding and referring capabilities through multi-task learning, and MAIRA-2 \cite{maira} enhances grounded report generation. Both approaches use textual coordinates to identify regions but rely on single-step inference and lack deeper reasoning capabilities for comprehensive clinical perception.
Med-ST\cite{yang2024unlocking} introduces modality-weighted local alignment between text tokens and spatial
regions of images.
fVLM \cite{shui2025large}
identifies false negatives of both normal
and abnormal samples and calibrating contrastive learning from patient-level to
disease-aware pairing.
MLMMs \cite{li2025aor} introduce a new dataset where chest X-ray (CXR) anatomical regions are represented as objects with attributes selected from 68 candidates across five categories, structured by three ontologies: object hierarchies, attribute causality, and object–attribute constraints.
PIBD \cite {zhangprototypical} learns prototypes to approximate a bunch of instances at different risk levels.
\cite{chen2023fine} introduces an adaptive patch extraction module to acquire adaptive patches on abnormal regions.
UMedLVLM \cite{zhou2025training} collects a Medical Abnormalities Unveiling (MAU) dataset to specifically annotate abnormal area.
Malenia \cite{jiang2024unleashing} leverages multiscale mask representations with inherent boundary information to capture diverse lesion regions, then
matches fine-grained visual features of lesions with text embedding.

From the \textbf{text level} perspective, recent work has focused on enhancing the semantic fidelity of textual representations.
\underline{\textit{Fine-Grained Contextual Understanding}} focuses on capturing the nuanced structure and semantics of medical narratives to support accurate and interpretable model predictions.
Models like MedKLIP \cite{wu2023medklip} and KAD \cite{zhang2023knowledge} enhance the text encoder with medical domain knowledge. Imitate \cite{liu2023imitate} introduces hierarchical alignment by separately aligning multi-level visual features with descriptive and conclusive segments of radiology reports, offering a more structured understanding of medical observations.
HSCR \cite{jiang2025hscr} proposes Hierarchical Self-Contrastive Reward to generate high-quality token-level responses. 
StructuralGLIP \cite{yangprompt} encodes prompts into a latent knowledge bank, enabling more
context-aware and fine-grained alignment.
\cite{liuinterpretable} considers the intrinsic ordinality knowledge in survival risks,
and designs ordinal survival prompts to encode continuous time-to-event information for pathology risk prediction.
TabPedia \cite{zhao2024tabpedia} learns a hierarchy of four tasks, including table detection, structure recognition, table querying, and question answering, to align table-centric images with textual information at varying levels of visual-semantic granularity.

\subsubsection{Data-Efficient Parallel and Unpaired Alignment}
Multimodal conversational AI has advanced rapidly through large-scale pretraining on billions of image-text pairs from the public web. However, acquiring sufficient high-quality image-text pairs in the biomedical domain remains a major bottleneck. Existing efforts to address this challenge can be broadly categorized into two strategies: (1) \textbf{constructing parallel biomedical datasets} for instruction-tuning, and (2) \textbf{learning from unpaired or weakly aligned data} through contrastive or generative objectives.

\underline{\textit{Parallel Data Collection and Construction}}.
LVM-Med \cite{mh2023lvm} aggregates 1.3 million medical images from 55 public datasets across various organs and modalities (CT, MRI, X-ray, Ultrasound), and introduces a novel graph-based contrastive learning algorithm for medical representation learning, alongside benchmarking existing self-supervised methods.
HealthGPT \cite{lin2025healthgpt} progressively adapts a pre-trained LLM as an unified medical VLM with a small amount of visual instruction data.
Biomedclip \cite{zhang2023biomedclip} constructed PMC-15M, a large-scale parallel dataset containing 15 million figure-caption pairs extracted from 4.4 million biomedical articles in PubMed Central. Building on this, LLaVA-Med \cite{li2023llava} introduces a novel data generation pipeline that samples image-text pairs from PMC-15M and uses GPT-4 for self-instruction to generate instruction-following data. The resulting dataset is then used to fine-tune a general-domain vision-language model for answering open-ended biomedical image questions.
Unlocking from the requirements of paired data preparation, Med-Flamingo \cite{moor2023med} continue pre-training on paired and interleaved
medical image-text data from publications and
textbooks based on OpenFlamingo-9B \cite{awadalla2023openflamingo}.

\underline{\textit{Learning from unpaired Data}}.
MedCLIP \cite{wang2022medclip} and PTUnifier \cite{chen2023towards} rely on unpaired image and report corpora, while PairAug \cite{xie2024pairaug} proposes pairwise augmentation by manipulating existing pairs or synthesizing new ones.
Med-MLLM \cite{liu2023medical} learns generalizable radiograph and clinical representations from unlabeled visual and textual data, enabling rapid adaptation to rare diseases with minimal labeled examples for diagnosis, prognosis, and report generation.
M3Care \cite{zhang2022m3care} imputes task-relevant information for missing modalities in the latent space using auxiliary data from clinically similar patients.
Muse \cite{wu2024multimodal} uses a flexible bipartite graph to represent
the patient-modality relationship, which can adapt to various missing modality
patterns.

\subsection{Temporal Modeling in Medical Vision-Language Model}
Existing works predominantly assume a static setting in which medical images, associated EHRs, and other modalities are naturally time-aligned. However, in real-world scenarios, these modalities are often highly asynchronous due to factors such as collection delays, posing significant challenges for effective multimodal information integration.

DDL-CXR \cite{yao2024addressing} dynamically generates up-to-date latent representations of CXR images to address modality asynchrony.
\cite{bannur2023learning} proposes a multi-image encoder that handles missing image inputs and incorporates longitudinal information without requiring explicit image registration.
\cite{liuinterpretable} encodes textual prognostic prior into prompts and then employs it as
auxiliary signals to guide the aggregating of visual prognostic features at instance level, thereby compensating for the weak supervision for risk prediction.
Med-ST \cite{yang2024unlocking} effectively incorporates temporal structure by modeling longitudinal dependencies through cycle-consistent learning objectives, significantly improving the temporal reasoning capability of medical VLMs.
Time-to-event pretraining \cite{huo2025timeto} introduces a longitudinally-supervised pretraining framework that aligns 3D medical imaging with temporal outcomes derived from EHRs
\cite{zhang2023improving} introduces an interleaved attention mechanism across time steps to model irregularity in multimodal fusion.

\subsection{Learning from External knowledge}

We categorize medical knowledge injection strategies in vision-language models based on the source of knowledge: (1) \textbf{external databases}, (2) \textbf{internalized knowledge within LLMs}, and (3) \textbf{case-based knowledge from similar patients}. These strategies are often implemented via prompt design or instruction-tuning data construction.

\underline{\textit{Learning from External Database}}.
AOR \cite{li2025aor} introduces anatomical ontologies to create a structured instruction-tuning dataset, enabling region-aware, interpretable, and multi-step reasoning over chest X-rays. Similarly, \cite{liu2024zero} incorporates clinician-verified knowledge from external databases into test-time prompts to reduce hallucination and enhance zero-shot classification.
StructuralGLIP \cite{yangprompt} introduces disease category-level descriptions into prompts to enable more
context-aware and fine-grained alignment.
\underline{\textit{Internalized Knowledge from LLMs}}.
EMERGE \cite{zhu2024emerge} leverages the implicit medical knowledge encoded in LLMs to generate comprehensive summaries of patients’ clinical states, which are then integrated into downstream decision-making tasks.
LLM-CXR \cite{lee2024llm-cxr} teaches the model visual knowledge by prompting it with diverse instructions related to CXR image, and then fine-tuning the LLM using the generated outputs.
\underline{\textit{Case-Based Knowledge from Patient Records}}.
Re3Writer \cite{liu2022retrieve} emulates physician workflow by retrieving similar historical patient instructions, reasoning over a learned medical knowledge graph, and refining this information to generate personalized discharge summaries.

\section{LLM-Based Modeling and Systems }
\label{sec::llms}

\input{Figures/section71/figure_section71}
\subsection{Prompt Engineering Approaches}

\subsubsection{Zero/few-shot Prompting}
DearLLM \cite{xu2025dearllm} leverages LLMs to infer personalized feature correlations in EHRs, enhancing medical predictions through patient-specific knowledge integration and graph-based modeling.
FeatLLM \cite{han2024featllm} employs LLMs as feature engineers to produce an input data set that is optimally suited for tabular
predictions. For example, LLMs can directly infer and generate decision rules that map specific feature conditions to diagnostic outcomes.
CAAFE \cite{hollmann2023large} utilizes LLMs to iteratively generate additional semantically meaningful features for tabular datasets based on the description of the dataset.
It generates both Python code for feature creation and natural language explanations that justify the utility of each generated feature.
HeartLang \cite{jin2025hegta} proposes a self-supervised framework that models ECG signals as language, using a QRS-Tokenizer to convert heartbeats into semantically meaningful sequences and learning form- and rhythm-level representations.
\cite{alsentzer2023zero} demonstrates that Flan-T5, a publicly available LLM, effectively extracts 24 granular PPH-related concepts from clinical notes without requiring annotated data.
\cite{swaminathan2023natural} trains on 721 clinical chats (32$\%$ crises) and proposes two-stage NLP system (keyword filtering + logistic regression) to prioritize crisis messages (e.g., suicidal ideation, domestic violence) in telehealth platforms, enabling faster triage and response during mental health emergencies.

Existing methods primarily leverage LLMs for feature transformation by uncovering informative correlations, inducing interpretable rules, and generating semantically enriched features, thereby improving predictive performance and interpretability. 
However, these works often overlook the heterogeneous nature of EHR data and lack explicit mechanisms to interpret numerical values within their clinical context.

\subsubsection{Structured Prompting}
CoK \cite{lichain} augments LLMs by dynamically incorporating grounding information
from heterogeneous sources. 
It first generates preliminary rationales, identifies relevant knowledge domains, and iteratively refines reasoning through domain-specific information integration.
ChAIN-OF-TABLE \cite{wangchain} introduces a tabular reasoning framework where tables serve as intermediate representations of thought. Using in-context learning, LLMs iteratively generate operations and update the table, enabling dynamic reasoning by planning each step based on previous outcomes.
Trompt \cite{chen2023trompt} is a prompt-based framework for deriving sample-specific feature importances. It comprises multiple Trompt Cells for feature extraction and a shared Trompt Downstream module for prediction.

Structured prompting methods enhance LLM reasoning by introducing intermediate representations, such as rationales, tabular states, or modular prompts, to decompose complex tasks. These approaches improve traceability, allow iterative refinement, and enable sample-specific adaptations. However, they often rely on handcrafted designs or predefined modules, which may limit scalability and generalization across diverse clinical scenarios. Integrating domain-adaptive prompting with structured semantics remains an open challenge for robust EHR modeling.

\subsection{Pretraining and Fine-tuning Methods}
MedFound \cite{liu2025generalist} is a 176-billion-parameter medical large language model pretrained on diverse medical literature and 8.7 million real-world electronic health records to embed domain-specific clinical knowledge. 
Me-LLaMA \cite{xie2025medical} is a family of open-source medical LLMs developed via continual pretraining on both biomedical literature and clinical notes (129B tokens).  Me-LLaMA integrates domain-specific knowledge and instruction-following ability for broad generalization across six medical NLP tasks, including QA, NER, relation extraction, and summarization.
Med-PaLM2 \cite{singhal2025toward} is a large-scale medical language model built on PaLM 2, improved via domain-specific fine-tuning and innovative prompting techniques such as ensemble refinement and chain of retrieval. It achieves state-of-the-art results across multiple medical question-answering benchmarks, including USMLE-style exams, and shows comparable or superior performance to generalist physicians in real-world clinical consultations.
Meerkat \cite{kim2025small} introduces a family of on-premises medical small language models (7B/8B) fine-tuned with chain-of-thought (CoT) reasoning paths extracted from 18 medical textbooks and MedQA-style questions. 
MOTOR \cite{steinbergmotor} is a self-supervised foundation model for time-to-event (TTE) prediction, pretrained on timestamped sequences from 55M patient records.
UniTabE \cite{yang2024unitabe} represents each table element using a modular unit (TabUnit), followed by a Transformer encoder for contextual refinement. The model supports both pretraining and finetuning via free-form prompts, enabling flexible adaptation to diverse tabular tasks. \cite{abbaspourazad2024wearable} introduces the first large-scale foundation models for PPG and ECG biosignals, pretrained via a self-supervised framework combining domain-specific and vision-inspired techniques on data from 141,207 wearable-device users, and demonstrates that the learned embeddings encode rich demographic and clinical information.
TP-BERTa \cite{yan2024making} is a pretrained language model tailored for tabular data prediction, introducing relative magnitude tokenization for encoding numerical values and intra-feature attention to jointly model feature names and values, enabling effective knowledge transfer across heterogeneous tables.
AMIE \cite{mcduff2025towards} is a large language model optimized for diagnostic reasoning, evaluated for its ability to generate differential diagnoses both independently and as an assistive tool to enhance clinician decision-making in complex medical cases.
MERA \cite{ma2025memorize} is a LLMs for clinical diagnosis prediction that models patient histories as sequences and predicts future diagnoses by optimizing outcome-level probabilities, incorporating hierarchical contrastive learning over ICD codes, intra-visit code ranking via teacher forcing, and semantic alignment through fine-tuning on code-definition mappings.
Med-PaLM 2 \cite{singhal2025toward} uses a combination of an improved base LLM, medical domain-specific fine-tuning and new prompting strategies to improve reasoning and grounding, including ensemble refinement and chain of retrieval.
\cite{chen2023parameterizing} introduces a continual table semantic parser that integrates parameter-efficient fine-tuning (PEFT) and in-context tuning (ICT) via a teacher–student framework, where the teacher uses ICT for few-shot learning and the student distills contextual knowledge into compact prompts, effectively mitigating catastrophic forgetting without storing training examples.
GatorTron \cite{yang2022large} uses more than 90 billion words of text (including >82 billion words of de-identified clinical text) and systematically evaluate it on five clinical NLP tasks including clinical concept extraction, medical relation extraction, semantic textual similarity, natural language inference (NLI), and medical question answering (MQA).

\label{sec::Rag}
\subsection{Retrieval-Augmented Generation}
TableRAG \cite{chen2024tablerag} introduces a RAG framework for table understanding, combining query expansion with schema and cell retrieval to supply LLMs with focused, high-utility tabular context, enabling more efficient prompting and reducing information loss in long-table queries.
OpenTable \cite{kongopentab} combines table retrieval with program synthesis by first retrieving relevant tables and then generating SQL queries to parse them. It uses the intermediate results from SQL execution to perform grounded reasoning, enabling accurate and context-aware response generation.
Re$^{3}$Writer \cite{liu2022retrieve}, mimics physicians’ discharge workflows by analyzing a patient’s diagnoses, medications, and procedures to automatically generate personalized post-discharge instructions (PI) using structured health records.
However, its performance depends heavily on the quality of retrieval and coverage of the medical knowledge graph, which may limit generalization in rare or complex cases.

\cite{kresevic2024optimization} integrates clinical guidelines via retrieval-augmented generation (RAG), prompt engineering, and text reformatting to improve guideline-grounded clinical decision support. Applied to hepatitis C virus (HCV) management, the model significantly outperforms baseline LLMs in generating accurate, guideline-aligned recommendations, as validated by expert review and text similarity metrics.
\cite{wang2024prompt} investigates how prompt engineering influences LLM alignment with evidence-based clinical guidelines, using AAOS osteoarthritis recommendations as a benchmark. By querying multiple LLMs with varied prompts and measuring consistency across five trials per question, the study finds that prompt effects vary by model. GPT-4-Web with ROT prompting achieved the highest consistency, highlighting the potential of tailored prompts to enhance LLM accuracy in medical question answering.
The Next Generation Evidence (NGE) system \cite{borchert2025high} leverages NLP to extract structured data from clinical trials and guidelines, enabling high-precision signal detection for timely guideline updates. By mapping content to UMLS concepts and supporting multilingual inputs, NGE reduces evidence synthesis delays and enhances decision-making for guideline developers and clinicians.
\cite{ke2025retrieval} proposes an LLM-RAG pipeline for preoperative medicine, aiming to reduce last-minute surgery cancellations by providing guideline-grounded triage and personalized instructions. By embedding institutional perioperative protocols into retrieval-augmented generation, the system improves decision support on patient fitness, fasting, and medication management, offering a scalable alternative to costly manual evaluations.

\subsection{LLM-Driven Medical Agents}
\input{Tables/comparison_agent}
Medical agents operate through three fundamental and sequential stages to accomplish clinical tasks. First, \textbf{perception} processes multimodal clinical inputs, such as unstructured text, medical images, and structured EHR data, to extract meaningful information. Second, \textbf{planning and reasoning} integrate clinical knowledge and logical frameworks to determine appropriate diagnostic or therapeutic strategies. Third, \textbf{execution} interfaces with external tools or environments to retrieve evidence, generate reports, or deliver actionable recommendations. Enabled by advances in LLMs, clinical agents are increasingly capable of goal-directed reasoning and decision-making in complex medical contexts. These agents aim to emulate key aspects of human workflows, including sequential diagnosis, multidisciplinary reasoning, and personalized care planning. 

\subsubsection{Memory Mechanism}.
Memory mechanisms enable agents to store, organize, and retrieve information across different temporal scopes, thereby enhancing contextual reasoning and decision-making. 
Depending
on its source, memory can generally be divided into 1) \textbf{Short-term
Memory}, 2) \textbf{Long-term Memory} and 3) \textbf{Knowledge Retrieval as Memory}.
\underline{\textit{Short-term memory}} retains internal dialog history and immediate environmental feedback to support context-sensitive task execution.
For instance, EHR2Path \cite{pellegrini2025ehrs} encodes patient histories into topic-specific summary tokens to preserve temporal context, while DiagnosisGPT \cite{chen2024cod} models the diagnostic process as a stepwise chain of reasoning that mirrors a physician’s cognitive workflow.
\underline{\textit{Long-term memory}} captures intermediate reasoning trajectories and transforms them into reusable knowledge assets that enhance future task performance. 
Current implementations exhibit two dominant approaches: 1) Experience repositories. For example,  EHRAgent \cite{shi2024ehragent} maintains a dynamic repository of successful execution cases for few-shot prompting and iterative refinement.
2) Tool synthesis. Agents may compose or reconfigure tools based on learned reasoning patterns, though this remains underexplored in clinical applications.
\underline{\textit{Knowledge retrieval as memory}} extends internal memory with access to external knowledge sources, expanding the agent’s information capacity and enabling real-time augmentation. 
This paradigm can be categorized into three themes.
\textit{Static knowledge grounding} retrieves structured information from fixed corpora such as clinical guidelines or medical ontologies. For example, ColaCare \cite{wang2025colacare} and MedAgent-Pro \cite{wang2025medagent}  employ retrieval-augmented generation to align decisions with domain knowledge. 
\textit{Interactive retrieval} dynamically queries external sources during multi-turn reasoning to adapt to evolving context, although it remains underexplored in clinical agents. 
\textit{Reasoning-integrated retrieval} represents an emerging paradigm where retrieval is interleaved with planning in a closed-loop fashion; however, current implementations in clinical domains are limited.

\subsubsection{Planning and Reasoning}
Planning enables clinical agents to decompose complex tasks, coordinate tool usage, and structure reasoning across time. 
Existing studies can be generally categorized into three
classes: 
1) \textbf{Static Planning},
2) \textbf{Dynamic planning},
and 3) \textbf{Iterative Planning} .

\underline{\textit{Static Planning}} relies on predefined, non-adaptive action sequences. These workflows typically prioritize structure and interpretability over flexibility. For example, 
EHR2Path \cite{pellegrini2025ehrs} encodes patient trajectories using topic-specific summary tokens and applies a fixed model for next-step prediction without real-time decision adaptation.
DiagnosisGPT \cite{chen2024cod} emulates a physician’s diagnostic chain with fixed steps and outputs a confidence distribution, enhancing transparency but lacking dynamic tool invocation.
CT-Agent \cite{mao2025ct} and MedRAX \cite{fallahpour2025medrax} integrate static expert modules to handle multimodal inputs (e.g., CT and CXR) in fixed pipelines.
\underline{\textit{Dynamic planning}} allows agents to compose task-specific action plans and tool selections based on current inputs.
For example, ColaCare \cite{wang2025colacare} adopts a meta-agent architecture that dynamically orchestrates consultations among domain-specific agents, reflecting clinical MDT processes.
MedAgent-Pro \cite{wang2025medagent} separates reasoning into disease-level guideline-grounded planning and patient-level adaptive reasoning steps.
MMedAgent \cite{li2024mmedagent} selects tools (e.g., grounding, segmentation, retrieval) on-the-fly using instruction-based learning across imaging modalities.
EHRAgent \cite{shi2024ehragent} translates EHR question-answering into dynamic tool-use sequences via code generation conditioned on query context and data structure.
\underline{\textit{Iterative Planning}} incorporates feedback from execution environments or tools to refine multi-step reasoning.
For example, MedAgent-Pro \cite{wang2025medagent} validates intermediate conclusions and revises plans via evidence-grounded checks.
EHRAgent \cite{shi2024ehragent} uses a multi-turn interactive coding loop where the agent iteratively revises plans based on code execution and error messages.

\subsubsection{Action execution} Action execution refers to an agent’s ability to interface with external environments or tools to carry out planned decisions. This capability is central to grounding abstract reasoning into practical outcomes. We categorize execution into three modes: 
(1) \textbf{Tool-Based Execution,}
(2) \textbf{Code-Based Execution}
and (3) \textbf{Environment-Embedded Execution}.

\underline{\textit{Tool-Based Execution}}.
Agents rely on domain-specific models or pretrained components as tools to perform discrete clinical actions. For example, ColaCare \cite{wang2025colacare} leverages expert models (e.g., mortality and readmission predictors) as DoctorAgents. 
MedAgent-Pro \cite{wang2025medagent} invokes visual models and retrieval tools to support stepwise reasoning grounded in clinical evidence. 
MMedAgent \cite{li2024mmedagent} orchestrates toolchains for multimodal tasks by aligning user instructions with task-specific tool execution.
MedRAX \cite{fallahpour2025medrax} integrates CXR expert modules and orchestrates their execution through a multimodal LLM interface.
(2) \underline{\textit{Code-Based Execution.}}
Instead of predefined tools, agents synthesize and execute code to solve tasks flexibly. EHRAgent formulates EHR reasoning as an executable code plan and interacts with a code executor to iteratively debug and improve task completion. This coding-based approach enables compositional operations across heterogeneous EHR tables with minimal supervision.
(3) \underline{\textit{Environment-Embedded Execution.}}
Agents operate within simulated or real clinical environments where action involves environmental manipulation, decision logging, or simulated patient interaction. MedAgentBench \cite{jiang2025medagentbench} evaluates execution within a FHIR-compliant virtual EHR system, while AgentClinic \cite{schmidgall2024agentclinic} embeds agents in patient simulations across specialties, requiring sequential decisions such as test selection, note-taking, and treatment planning.

\subsubsection{Self-Improvement} \textit{Self-Improvement} refers to an agent’s ability to iteratively enhance its performance by learning from past experiences, execution feedback, or external evaluation. We categorize this capability into three paradigms: 
(1) \textbf{Memory-Based Self-Improvement}, 
(2) \textbf{Interactive Refinement}, 
(3) \textbf{Tool Adaptation and Extension}.
\underline{\textit{Memory-Based Self-Improvement}}. Agents equipped with long-term memory can accumulate and reuse experience over time. 
EHRAgent \cite{shi2024ehragent} maintains a dynamic repository of successful code plans and iteratively selects the most relevant few-shot examples to guide new tasks, enabling progressive refinement across EHR reasoning problems. 
(2) \underline{\textit{Interactive Refinement}}.
Agents refine their internal reasoning by incorporating feedback from tool execution, error messages, or multi-turn planning.
For example, 
MedAgent-Pro \cite{wang2025medagent} enforces step-by-step evidence verification at each diagnostic stage. If uncertainty or contradictions are detected, the agent re-evaluates previous steps using retrieved guideline evidence.
DiagnosisGPT \cite{chen2024cod} improves interpretability and control by analyzing entropy across confidence distributions, identifying symptoms with highest uncertainty, and prompting targeted clarification.
(3) \underline{\textit{Tool Adaptation and Extension}}.
Agents that dynamically update or integrate tools during deployment exhibit another form of self-improvement. 
For example, 
MMedAgent \cite{li2024mmedagent} incorporates new medical tools (e.g., for grounding or segmentation) and learns to select them appropriately via instruction tuning. Its architecture supports efficient tool updates without retraining the entire model.
ColaCare \cite{wang2025colacare} demonstrates adaptability by allowing its LLM-driven agents to interact with evolving structured expert models and update decision-making with new clinical guidelines via RAG.
AgentClinic \cite{schmidgall2024agentclinic} tracks agent behavior under incomplete information and simulates patient feedback. It supports experiential learning by storing previous case notes and examining diagnostic accuracy shifts across iterations.

\subsubsection{Multi-Agent Collaborations}
Collaboration mechanisms allow clinical agents to divide responsibilities, communicate across specialized roles, and reason collectively toward shared clinical goals. Inspired by real-world multidisciplinary team (MDT) practices, agent collaboration in current systems can be categorized into three forms : 
(1) \textbf{Centralized Control},
(2) \textbf{Decentralized collaboration},
and (3) \textbf{Hybrid architectures},
In \underline{\textit{centralized control}} systems, a single agent (often called a meta-agent or manager) oversees the execution of subtasks by subordinate agents or modules.
ColaCare \cite{wang2025colacare} exemplifies centralized control through a MetaAgent that coordinates multiple DoctorAgents, each representing a domain-specific expert. The MetaAgent manages task assignment, evidence aggregation, and final decision formulation.
MedAgent-Pro \cite{wang2025medagent} implicitly follows a centralized structure by planning at the disease level and then orchestrating patient-level reasoning with tool invocations, guided by retrieved guideline evidence.
(2) \underline{\textit{Decentralized collaboration}} involves multiple autonomous agents contributing to a shared task without a controlling agent. Each agent operates based on local observations and exchanges information directly.
While not yet common in clinical LLM systems, multi-agent dialog frameworks (e.g., inspired by Chain-of-Agents or swarm-agent settings) may evolve toward this design. 
(3) \underline{\textit{Hybrid architectures}} blend centralized coordination with decentralized flexibility. For example, a manager may initiate tasks, but agents can exchange information and adapt dynamically during execution.
MMedAgent \cite{li2024mmedagent} hints at a hybrid model by coordinating tool-based agents (e.g., for segmentation, grounding) under an instruction-following framework. Although execution is managed centrally, tools are modular and extensible, enabling agent-like behavior from components.

%% file: Tables/Multimodality.tex
\begin{table}[ht]
\centering
\resizebox{1.0\textwidth}{!}{ 
\begin{tabular}{@{}p{3.2cm}p{4.5cm}p{4.5cm}p{4.5cm}p{4.5cm}@{}}
\toprule
\textbf{Category} 
& \textbf{Computed Tomography (CT)} 
& \textbf{Magnetic Resonance Imaging (MRI)} 
& \textbf{Digital Radiography (DR)} 
& \textbf{Ultrasound (US)} 
\\
\midrule
Overview 
& CT is a process using X-rays to rapidly scan the body in slices and generate 3D structural images. 
& MRI uses strong magnetic fields and radio waves passing through the body. Protons inside respond to generate detailed structural images. 
& DR is a digital form of X-ray examination that produces digital X-ray images instantly on a computer. 
& Ultrasound uses high-frequency sound waves to observe internal conditions. It provides real-time images of organs and blood flow changes.
\\
\midrule
Average Scan Duration 
& Approx. 5 minutes 
& 10–30 minutes 
& Approx. 2 minutes 
& N.A. 
\\
\midrule
Radiation 
& \ding{51} 
& \ding{55} 
& \ding{51} 
& \ding{55} 
\\
\midrule
Soft Tissue Imaging 
& \ding{51} 
& \ding{51} 
& \ding{55} 
& \ding{51} 
\\
\midrule
Main Imaging Areas 
& Orthopedics, Brain, Lungs, Cardiology 
& Liver, Soft Tissue, Central Nervous System, Musculoskeletal 
& Orthopedics, Gastroenterology, Intestines, Respiratory System 
& Orthopedics, Gastroenterology, Intestines, Respiratory System 
\\
\midrule
Image Resolution 
& High 
& Highest 
& Medium 
& Medium 
\\
\bottomrule
\end{tabular}
}
\small
\raggedright
\caption{Comparison of Medical Imaging Modalities}
\end{table}
\vspace{-1cm}

%% file: Figures/section71/figure_section71.tex
\begin{figure}[htbp]
  \centering
  \includegraphics[width=1\textwidth]{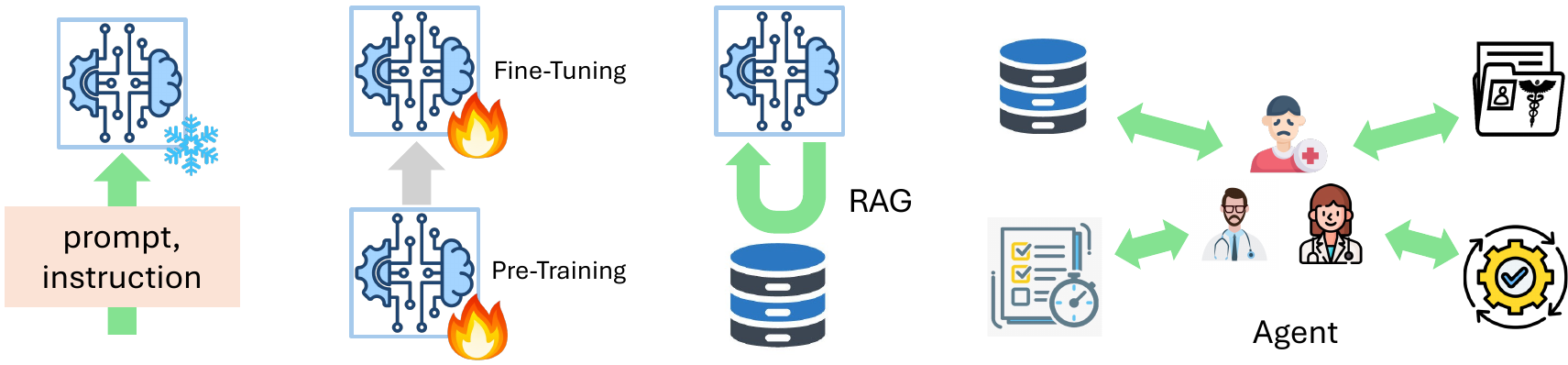}
  \caption{An illustration of learning with large language model strategies.}
  \label{LLM methods}
\end{figure}

%% file: Tables/comparison_agent.tex
\begin{table}[ht]
\centering
\resizebox{1.0\textwidth}{!}{ 
\begin{tabular}{l|c|c|c|c|c}
\hline
\textbf{Paper} & \textbf{Memory} & \textbf{Planning} & \textbf{Action Execution} & \textbf{Collaboration} & \textbf{Self-Improvement} \\
\hline
\textbf{EHR2Path}       & \checkmark &  &  &  &  \\
\textbf{ColaCare}       & \checkmark & \checkmark & \checkmark & \checkmark &  \\
\textbf{CT-Agent}       & \checkmark & \checkmark & \checkmark &  &  \\
\textbf{MedRAX}         & \checkmark &  & \checkmark &  &  \\
\textbf{MedAgentBench}  &  & \checkmark & \checkmark &  &  \\
\textbf{MedAgent-Pro}   & \checkmark & \checkmark & \checkmark &  & \checkmark \\
\textbf{DiagnosisGPT} & \checkmark & \checkmark &  &  & \checkmark \\
\textbf{MMedAgent}      &  & \checkmark & \checkmark &  & \checkmark \\
\textbf{EHRAgent}       & \checkmark & \checkmark & \checkmark &  & \checkmark \\
\textbf{AgentClinic}    & \checkmark & \checkmark & \checkmark &  & \checkmark \\
\hline
\end{tabular}
}
\caption{Comparison of existing clinical agents.}
\label{tab::agents}
\end{table}

%% file: Sections/Methodology/downstream_tasks.tex
\section{Clinical Applications}
\label{sec::application}
\input{Figures/section8/figure_section8}
\input{Tables/downstream}
\subsection{Clinical Document Understanding}
\subsubsection{Clinical Document Summarization}
\textbf{Clinical documents} capture both narrative and structured representations of patient care, including the history of present illness, physical exam findings, diagnostic results, clinical assessments, treatment plans, and discharge summaries. These records are critical for care continuity, billing, legal documentation, and provider communication \cite{yackel2010unintended,bowman2013impact}.
\textbf{Clinical document summarization} seeks to condense lengthy or unstructured narratives into concise, coherent summaries to support downstream tasks such as patient handoffs, triage, retrospective review, and decision support \cite{ghosh2024sights}. The task presents two major challenges: (1) preserving factual accuracy to avoid clinical misinformation, and (2) handling heterogeneous input formats across note types and specialties.

Existing work on clinical document summarization can be broadly categorized into two directions:
(i) \textbf{Extractive approaches} aim to identify salient sentences or phrases from the original document while preserving their exact wording \cite{wang2020cord} . 
One line of work focuses on leveraging attention mechanisms \cite{deyoung2021ms2} or supervised relevance scoring \cite{wallace2021generating} to highlight key content from clinical notes or discharge summaries. Another direction incorporates document structural information, first extracting layout or section formatting and then identifying informative content within those structures~\cite{joshi2020dr, poornash2023aptsumm}. These approaches are generally more reliable for preserving factual accuracy and are particularly suited for high-stakes clinical summarization.
(ii) \textbf{Abstractive approaches}, which generate novel summaries by paraphrasing \cite{deyoung2021ms2} or synthesizing information \cite{wallace2021generating} from the input text \cite{wang2020cord}. 
Early works adopted sequence to sequence models such as BART~\cite{lewis2019bart}, PEGASUS~\cite{zhang2020pegasus}, and ProphetNet~\cite{qi2020prophetnet}. 
More recent studies have leveraged LLMs for clinical document summarization. Goff and Loehfelm~\cite{goff2018automated} applied pretrained models to summarize the impression sections of radiology reports, while Rajaganapathy et al.\cite{rajaganapathy2025synoptic} fine-tuned LLaMA 2 to generate structured synoptic pathology reports across multiple cancer types. Van Auken et al.\cite{van2024adapted} and Asgari et al.~\cite{asgari2025framework} proposed comprehensive evaluation frameworks, emphasizing factual consistency, clinical safety, and error categorization in LLM-generated summaries.

\subsubsection{Clinical Note Generation} Automatic generation of clinical documentation, including radiology reports, discharge summaries, and progress notes.
For example, \textit{radiology report generation} aims to automatically produce narrative or structured descriptions from medical images, mirroring the nuanced, context-rich communication of expert radiologists. These systems support a range of clinical and operational use cases, including assistive draft generation, report automation in resource-limited settings, radiology education via exemplar reports, and retrospective quality audits.
Flamingo-CXR \cite{tanno2025collaboration} adapts a pretrained vision–language foundation model using few-shot learning and domain-specific CXR–report pairs. It retains frozen visual and language backbones while fine-tuning cross-modal layers, enabling accurate radiology report generation without full supervised retraining. This strategy achieves state-of-the-art performance in both automated and expert evaluations.

\subsubsection{Concept Extraction}
\textbf{Concept Extraction} aims to identify clinically relevant entities such as diseases, symptoms, medications, and procedures from clinical text and map them to standardized medical vocabularies or ontologies, including UMLS, SNOMED CT, ICD-10, and RxNorm.
It enables downstream tasks such as clinical decision support, semantic search, cohort identification, clinical trial matching, and large-scale analytics by mapping free-text entities to controlled medical vocabularies. 
Extracting clinical concepts from free-text notes remains a challenging task due to the complexity of medical language. Common difficulties include ambiguous abbreviations, negation, temporal expressions, and fine-grained descriptions of symptom severity. 

Existing works can be categorized into three directions:
(1) \textbf{Rule-Based Methods}.
Early systems such as MetaMap \cite{aronson2010overview}, cTAKES \cite{savova2010mayo}, MedLEE \cite{friedman1994general}, and CLAMP \cite{soysal2018clamp} relied on manually crafted rules and dictionary lookups, typically leveraging the UMLS and other controlled vocabularies. These approaches suffers from limited scalability, poor recall, and poor generalizability across institutions and clinical specialties.
(2) \textbf{Embedding-Based Methods}.
With the rise of machine learning, research shifted toward learning semantic representations of clinical text using neural architectures. Tools such as ScispaCy \cite{neumann2019scispacy}, medspaCy \cite{eyre2022launching}, and EHRKit \cite{li2022ehrkit} incorporated contextual embeddings and syntactic features to improve robustness. Domain-specific transformer models like UmlsBERT \cite{michalopoulos2020umlsbert} and CancerBERT \cite{zhou2022cancerbert} were pretrained on biomedical corpora to capture fine-grained clinical semantics, significantly advancing concept recognition performance.
(3) \textbf{LLM-Based Methods}.
Recent advances leverage LLMs for more flexible, adaptive, and high-precision extraction.
For example,
CLEAR \cite{lopez2025clinical} introduces a clinically guided retrieval-augmented generation (RAG) pipeline, using named entities for chunk retrieval to improve extraction relevance and accuracy.
Keloth et al. \cite{keloth2025social} focus on Social Determinants of Health (SDoH), presenting a large annotated corpus and evaluating the cross-institutional generalizability of LLM-based models.
Hein et al. \cite{hein2025iterative} propose a pipeline for pathology report normalization combining prompt-based generation, tabular output structuring, and human-in-the-loop error correction, demonstrating the utility of LLMs in structured data extraction workflows.

\subsubsection{Image-Text Retrieval}
Image–text retrieval in EHRs supports critical downstream tasks by aligning visual modalities (e.g., radiology, pathology, dermatology) with corresponding clinical narratives. There are several important applications include:
(1) \textbf{Automated report generation}, where retrieved historical cases enhance the consistency and accuracy of imaging reports;
(2) \textbf{Similar case retrieval}, which facilitates diagnosis and differential reasoning by surfacing matched patient profiles;
(3) \textbf{Clinical decision support}, enabling contextual recommendations by retrieving relevant prior cases, treatments, or guidelines; and
(4) Clinical trial matching, where imaging phenotypes are linked to text-based eligibility criteria to identify potential candidates.
Existing studies primarily construct image–text pairs by collecting clinical images with natural language descriptions from medical literature \cite{kim2024transparent} or by manually annotating medical concepts \cite{huang2024critical}.
A comprehensive literature review can refer to Section \ref{sec::Rag}.


\subsubsection{Clinical Coding Automation} Clinical code automation aims to translate clinical text into standardized medical codes that support billing, reporting, epidemiological analysis, and clinical research.
Recent advances in ICD coding have explored structured modeling, weak supervision, and LLMs to improve performance and scalability. To incorporate domain-specific structure, several models integrate attention mechanisms and medical ontologies. Li et al. \cite{li2020icd} proposed a residual CNN with label attention, while Xie et al. \cite{xie2019ehr} utilized the ICD hierarchy in a graph neural network to model label dependencies. Cao et al. \cite{cao2020hypercore} embedded codes into a hyperbolic space to capture both hierarchical relations and co-occurrence patterns, and Yuan et al. \cite{yuan2022code} enhanced label-text alignment using synonym-aware attention. In parallel, weak supervision has emerged as a cost-effective strategy to alleviate the burden of manual annotation. Dong et al. \cite{dong2021rare} employed SemEHR to detect rare disease mentions via entity linking, while Gao et al. \cite{gao2022classifying} developed KeyClass, which automatically generates labels based on embedding similarities between n-gram keywords and code descriptions. More recently, Pretrained models have been applied to ICD coding. Pretrained models like BERT~ \cite{devlin2019bert} show limited utility due to input length constraints and domain mismatch~ \cite{pascual2021towards}. 
Liu et al. \cite{huang2022plm} addressed length limitations by chunking input text, but at the cost of coherence. 
KEPT \cite{yang2022knowledge} offers a more integrated solution using Longformer~ \cite{beltagy2020longformer}, contrastive learning \cite{lu2023towards} for code synonyms, and prompt-based decoding.
CliniCoCo \cite{gao2024optimising} proposes a novel human-in-the-loop (HITL) framework integrates coder feedback throughout the automated clinical coding process to improve accuracy and efficiency.

\subsubsection{Quantitative Clinical Calculations}
Quantitative clinical calculations are numerical procedures that apply mathematical formulas or models to structured clinical data to inform diagnosis, risk assessment, and therapeutic decision-making.
For example, perioperative risk scoring requires the integration of multiple clinical variables such as patient age, comorbidities, laboratory results, and surgical complexity into standardized formulas or scoring systems. Manual calculation by clinicians is often time-consuming and prone to error due to complex arithmetic, variability in guideline interpretation, and potential omission of critical variables.
Recent work \cite{goodell2025large} is the first to incorporate a functional clinical calculator into an LLM. It evaluated ChatGPT on 48 clinical calculation tasks using 212 vignettes and introduced a novel error classification to identify reasoning and arithmetic mistakes. 
 
\subsubsection{General Clinical AI Capabilities}
Recent studies leverage LLMs trained on extensive biomedical and clinical datasets to enable general-purpose capabilities across tasks such as concept extraction, clinical summarization, and image–text retrieval in pathology \cite{lu2024visual,chen2024towards}, echocardiography \cite{christensen2024vision}, radiologists \cite{yu2024heterogeneity}, dermatology \cite{groh2024deep}, etc.

\subsection{Clinical Reasoning and Decision Support}

Clinical reasoning and decision support systems aim to assist clinicians in interpreting patient data, forming diagnoses, anticipating outcomes, and selecting appropriate treatments. This section is organized into three core components. Diagnosis prediction focuses on identifying existing or emerging conditions, including early detection and differential diagnosis. Prognostic forecasting emphasizes estimating future outcomes such as disease progression, readmission, or mortality. Treatment modeling involves learning effective intervention strategies through tasks such as treatment effect estimation, personalized recommendation, and counterfactual simulation. 
\subsubsection{Diagnosis Prediction} Diagnosis Prediction involves machine learning models that infer the presence, stage, or subtype of a disease based on clinical data. 
We categorize these works into \textbf{EHR-only} and \textbf{multimodal approaches} based on \textbf{disease type}. Chronic conditions are often identifiable through clinical history, laboratory trends, and structured EHR data alone. In contrast, diseases such as cancer and dermatologic disorders typically require both EHR and imaging data for accurate detection, motivating the development of multimodal models.
\underline{\textit{EHR-only models}} are primarily used for early detection of conditions that manifest through longitudinal trends or clinical history, including chronic diseases (e.g., diabetes, heart failure), ICU deterioration (e.g., sepsis, acute kidney injury), and asymptomatic progression detectable via labs or notes (e.g., liver fibrosis, early CKD).
\underline{\textit{Multimodal models}} are primarily applied to oncologic and cardiovascular diseases \cite{ng2023prospective, cao2023large}, as well as conditions requiring imaging-guided decision-making \cite{thieme2023deep}, such as nodule assessment \cite{yao2025multimodal}, organ segmentation, or disease staging, where combining imaging, reports, and clinical context improves diagnostic accuracy and clinical relevance.

Specifically, \textbf{\textit{Differential Diagnosis}} refers to the systematic process of distinguishing a target condition from other diseases with overlapping clinical presentations. It is essential for ensuring diagnostic accuracy, guiding appropriate treatment decisions, and minimizing unnecessary interventions. Given the high degree of symptom similarity across many conditions, accurate differentiation is critical to avoid misdiagnosis and its associated clinical risks.
Recent research in AI-enabled differential diagnosis primarily focuses on how to effectively leverage diverse forms of medical knowledge to improve diagnostic reasoning, particularly in complex and rare disease contexts. One prominent direction involves multimodal knowledge integration, where heterogeneous inputs—such as imaging, demographics, medical history, and cognitive assessments—are fused to reduce diagnostic ambiguity. For example, Xue et al.~\cite{xue2024ai} propose a multimodal model for cognitive disorder subtyping that enhances accuracy through cross-modal representation learning.
A second line of work focuses on phenotype-based reasoning, extracting structured phenotypic information from EHRs to support rare disease diagnosis \cite{mao2025phenotype}. 
A third line of work leverages open-ended clinical cases to improve the diagnostic reasoning capabilities of LLMs in realistic scenarios \cite{fast2024autonomous}. 
Finally, researchers have begun exploring collaborative multi-agent reasoning frameworks to simulate multidisciplinary team (MDT) discussions. Chen et al.~\cite{chen2025enhancing} propose a multi-agent conversation (MAC) architecture in which multiple LLM agents reason interactively to generate differential diagnoses. 

\subsubsection{Prognostic Forecasting} Prognostic forecasting refers to the task of predicting a patient’s future clinical outcomes based on current and historical data. These outcomes may include disease progression, survival time, recovery trajectory, complication risks, or treatment response.

\textbf{Risk Prediction} aims to estimate the likelihood that a clinical event will occur within a defined time interval. 
This may involve forecasting disease onset or identifying potential adverse behaviors to support timely intervention and informed clinical decision making.
Recent advances in healthcare risk prediction have led to the development of diverse modeling paradigms designed to capture the temporal, semantic, and relational complexities of electronic health record (EHR) data. 
These paradigms can be categorized as follows:
(1) \underline{\textit{Time-aware models}} incorporate temporal dynamics into model design.
RETAIN \cite{choi2016retain} employs two recurrent neural networks operating in reverse time order and uses an attention mechanism to generate context vectors for interpretability. 
T-LSTM \cite{baytas2017patient} introduces a subspace decomposition of the cell memory, allowing time-aware decay that adjusts memory content based on elapsed time.
(2) \underline{\textit{Attention-based models}} employ attention or self-attention to enhance sequence modeling \cite{dai2024deep, garriga2022machine,luo2020hitanet}. 
MCA-RNN \cite{lee2018diagnosis} integrates conditional generative modeling with attention to capture EHR heterogeneity. 
SAnD \cite{song2018attend} uses masked self-attention with positional encoding and dilation to process long sequences, while ConCare \cite{ma2020concare} integrates time-aware attention with multi-head self-attention to improve patient representation learning. 
(3) \underline{\textit{Knowledge-enhanced models}} aim to improve prediction by incorporating structured domain knowledge, such as medical ontologies and similar patient trajectories. Examples include GRAM \cite{choi2017gram}, KAME \cite{ma2018kame}, and HAP \cite{zhang2020hierarchical}, which utilize ICD hierarchies. 
GRASP \cite{zhang2021grasp} clusters similar patients to transfer disease-related knowledge.
(4) Graph-based models are developed to learn dependencies between temporally ordered events or between patients and medical concepts \cite{chen2024predictive}.

\textbf{Readmission Prediction} aims to estimate the likelihood that a patient will experience an unplanned hospital readmission within a defined time frame following discharge, typically within 7, 30, or 90 days. High readmission rates are often viewed as indicators of poor care quality or premature discharge. As such, accurate prediction is critical for improving care continuity, reducing preventable hospitalizations, and optimizing resource utilization.
Readmission Prediction is commonly framed as a binary classification problem. In some cases, clinical notes are also incorporated. Predictive models help identify high-risk patients early, enabling timely interventions such as targeted discharge planning and post-acute follow-up.
Recent advances in readmission prediction focus on learning patient representations that capture temporal dynamics, handle missing data, and model complex clinical trajectories. AutoDP \cite{cui2024automated} applies multi-task learning to jointly predict multiple outcomes, including readmission, by sharing representations across related tasks to improve generalization. 
SMART \cite{yu2024smart}  introduces self-supervised learning methods that incorporate missingness-aware attention and latent-space reconstruction. 
Instruction tuning has also been explored \cite{wu2024instruction} to adapt LLMs for structured readmission prediction tasks, aligning model behavior with clinical objectives and improving short-term forecasting performance.

\textbf{Patient length of stay and mortality prediction}. 
Length of stay prediction aims to estimate the duration of a patient’s hospitalization. Accurate forecasts enable hospitals to manage bed occupancy, staffing, and discharge coordination. 
Existing works mainly aim to model temporal dynamics and multimodal signals from EHRs. 
For example, DeepMPM \cite{yang2022deepmpm} introduces a two-level attention-based LSTM to capture both temporal and variable interactions in longitudinal EHR data, enabling interpretable in-hospital mortality prediction. 
TECO \cite{rong2025deep} proposes a transformer-based model that incorporates static and time-dependent features for ICU mortality prediction, demonstrating generalizability across sepsis, COVID-19, and ARDS cohorts.
ChronoFormer \cite{zhang2025chronoformer} explicit temporal modeling to improve performance on time-sensitive clinical tasks, including mortality forecasting. 


\subsubsection{Cohort Discovery and  Patient Stratification}   Cohort discovery and patient stratification are sequential and complementary tasks in clinical research and AI-driven healthcare. Cohort discovery identifies a relevant patient population based on predefined criteria, while patient stratification segments that cohort into subgroups with distinct clinical profiles or risks to enable more personalized analysis or intervention.

\textbf{Cohort Discovery} is the process of identifying a group of patients from a clinical database who meet a predefined set of inclusion and exclusion criteria. It is commonly used in clinical research, trial recruitment, and retrospective studies to assemble patient populations with shared characteristics such as diagnoses, treatments, demographics, or laboratory findings.
COMET \cite{mataraso2025machine} utilizes large-scale EHRs via transfer learning to augment the analysis of small-cohort multi-omics studies. By integrating pretrained EHR representations with multi-omics data, COMET enhances predictive performance and biological interpretability, effectively addressing challenges posed by high-dimensional data and limited sample sizes.

\textbf{Subtyping/Endotyping}. Patient stratification refers to the task of dividing patients into clinically meaningful subgroups based on characteristics such as disease severity, risk of adverse outcomes, or predicted treatment response. It is used to enable personalized care, optimize clinical decision-making, and support targeted interventions or trial design.
Existing works on patient subtyping and endotyping primarily focus on two main directions. The first involves uncovering biologically or clinically meaningful subgroups to better characterize disease heterogeneity. For example, Raverdy et al. \cite{raverdy2024data} identify distinct MASLD subtypes with divergent cardiometabolic risks using unsupervised clustering, while Virchow\cite{vorontsov2024foundation}, a histopathology foundation model, enables robust tumor subtyping across cancer types, including rare variants. The second direction focuses on identifying model-specific performance disparities by stratifying patients based on where machine learning models fail. AFISP~\cite{subbaswamy2024data} exemplifies this by detecting underperforming subgroups in clinical ML systems and generating interpretable characterizations to support bias analysis and fairness-aware evaluation.

\subsection{Clinical Administration and Workflow Optimization}
\subsubsection{Triage and Prioritization} Triage and Prioritization aims to optimize clinical workflows by identifying high-risk patients for timely intervention, particularly in settings constrained by cost, personnel, or imaging resources. 
Existing research on triage and prioritization can be grouped into three main directions. The first focus on task-priority decomposition, where the triage process is modularized into predictive subtasks (e.g., baseline risk, masking assessment, and cancer suspicion) to enable targeted, interpretable decision-making. For example, AISmartDensity ~\cite{salim2024ai} combines three predictive modules to prioritize patients for supplemental MRI following negative mammograms. The second direction leverages external knowledge sources as a basis for triage prioritization. For example, Chang et al.\cite{chang2025continuous} propose a “metadata supply chain” that integrates clinical, genomic, and imaging information to enable continuous, context-aware prioritization in oncology. The third direction applies LLMs as triage evaluators. For example, Gaber et al.\cite{gaber2025evaluating} demonstrate that LLMs can assist in emergency severity indexing, diagnostic reasoning, and referral decisions, ultimately improving clinical accuracy and resource efficiency.

\subsubsection{Referral recommendation}
Referral recommendation systems aim to assist primary care physicians (PCPs) in identifying patients who require specialist consultation based on clinical risk profiles, diagnostic uncertainty, or guideline-driven thresholds. These systems are particularly valuable in resource-limited settings, where timely and accurate referrals can significantly impact patient outcomes and healthcare efficiency.
Existing works primarily focus on leveraging deep learning and LLMs to develop self-referral chatbots that streamline intake processes, reduce stigma, and improve access to care, particularly in mental health and primary care contexts \cite{habicht2024closing}.
For example, DeepDR-LLM \cite{li2024integrated} integrates a diabetic retinopathy image classifier with a fine-tuned medical language model to provide interpretable and guideline-based referral recommendations for ophthalmology follow-up. 
SSPEC \cite{wan2024outpatient} supports frontline administrative staff by learning from large-scale nurse–patient conversations. It enables realistic dialogue simulation and has demonstrated clinical utility in randomized controlled trials. 
Groh et al. \cite{groh2024deep} investigate physician and AI collaboration in diagnostic decision-making. Their findings show that the effectiveness of AI assistance varies with clinician expertise and population diversity, emphasizing the need for personalized and equitable AI integration in clinical workflows.

\subsubsection{Patient–Trial Matching}
Patient–trial matching aims to identify eligible individuals for clinical trials based on their EHR data and the often complex eligibility criteria described in trial protocols. Traditional approaches rely on structured data and handcrafted rules, which limit scalability and adaptability in real-world settings. Recent advancements leverage large language models (LLMs) to parse unstructured clinical text and automate matching.
For example, 
PRISM \cite{gupta2024prism}  introduces a pipeline that fine-tunes a domain-specific LLM (OncoLLM) to directly match unstructured patient notes with textual trial criteria. This design eliminates the need for intermediate rule-based processing and demonstrates high performance in oncology trial recruitment.
Other approach \cite{alkhoury2025enhancing} adopts a structure-then-match paradigm. A fine-tuned open-source LLM is used to extract and normalize genomic biomarkers into disjunctive normal form (DNF), allowing logical reasoning over eligibility rules. This structured representation enables accurate and scalable trial matching, especially for genomics-guided studies.


%% file: Figures/section8/figure_section8.tex
\begin{figure}[htbp]
  \centering
  \includegraphics[width=1\textwidth]{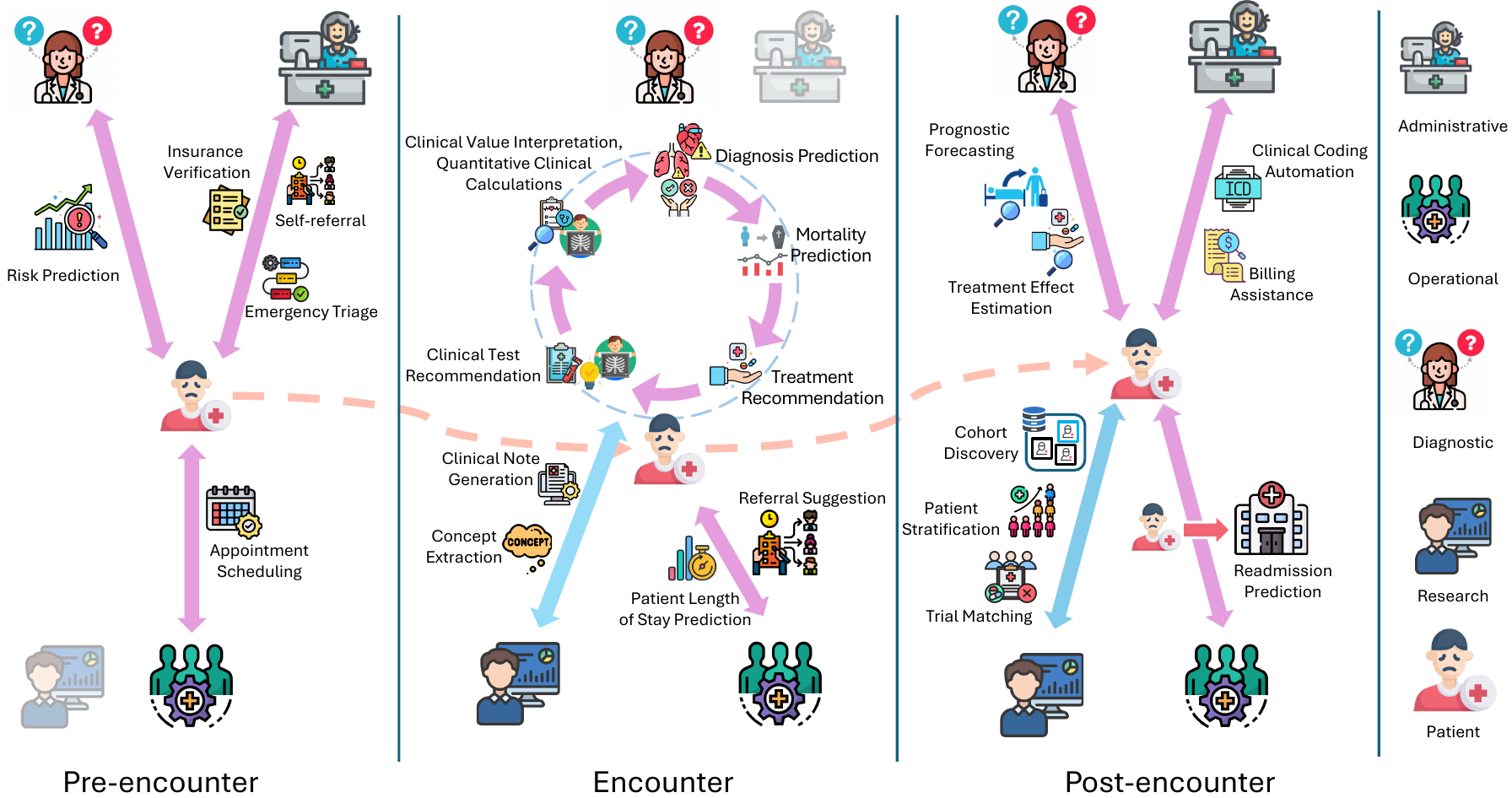}
  \caption{Illustration of the clinical decision support landscape across the care timeline. A typical patient trajectory can be categorized into three major phases: pre-encounter (e.g., risk prediction, appointment scheduling, emergency triage), encounter (e.g., diagnosis prediction, clinical test recommendation, treatment planning), and post-encounter (e.g., readmission prediction, cohort discovery, billing assistance). These stages can be further aligned with one or more functional roles, including administrative, operational, diagnostic, research, and patient-centered applications.}
  \label{app:workflow}
\end{figure}

%% file: Tables/downstream.tex
\begin{table}[ht]
\centering
\small
\begin{tabular}{c|c|c|c}
\hline
\textbf{Category} & \textbf{Pre-encounter} & \textbf{Encounter} & \textbf{Post-encounter} \\
\hline

\multirow{3}{*}{Administrative} 
& Insurance Verification &  & Clinical Coding Automation \\
& Self-referral &  & Billing Assistance \\
& Emergency Triage &  &  \\
\hline

\multirow{2}{*}{Operational} 
& Appointment Scheduling & Referral Suggestion & Readmission Prediction  \\
&  & Patient Length of Stay Prediction &  \\
\hline
\multirow{7}{*}{Diagnostic} 
& Risk Prediction & Clinical Test Recommendation &  Prognostic Forecasting\\
&  & Clinical Value Interpretation & Treatment Effect Estimation \\
&  & Quantitative Clinical Calculations &  \\
&  & Diagnosis Prediction &  \\
&  & Image–Text Retrieval &  \\
&&Mortality Prediction&\\
&&Treatment Recommendation&\\
\hline
\multirow{5}{*}{Research} 
&  & Clinical Note Generation & Cohort Discovery  \\
&  & Concept Extraction & Patient Stratification \\
&  &  & Trial Matching \\
\hline
\end{tabular}
\caption{Categorization of representative clinical AI tasks across the care timeline. Tasks are grouped into three phases: pre-encounter, encounter, and post-encounter. Each phase is further organized by functional categories, including administrative, operational, diagnostic, and research applications.}
\label{tab::function_taxonomy}
\end{table}

%% file: Sections/Experimental_design/exp.tex
\section{Evaluation Benchmark} 
\label{sec::dataset_evaluation}
In this section, we introduce publicly available
benchmark datasets with evaluation metrics according to the downstream tasks.
\cite{goh2025gpt} evaluates the potential of LLMs to augment physician clinical management decisions beyond diagnostic reasoning. Unlike traditional AI, LLMs can act as collaborative partners, integrating sequential clinical data to iteratively refine complex management plans. A prospective randomized controlled trial compares physicians aided by LLMs to those using standard resources and standalone LLM outputs on real patient cases, reflecting the dynamic and multifaceted nature of clinical decision-making. The work underscores the promise and challenges of applying LLMs to support nuanced treatment reasoning in healthcare.
Hailey \cite{sharma2023human} is an AI-in-the-loop agent that delivers real-time feedback to assist peer supporters in responding with greater empathy to individuals seeking help.

\subsection{Public EHR dataset}
Some of these datasets are single-purpose datasets for a particular downstream task,
and some are general-purpose time-series datasets that we can use for model evaluation across
different tasks. Table \ref{tab:dataset_stats} presents useful information of the reviewed datasets, including dataset names, sizes, diseases, reference sources, and application domains.

\subsubsection{MIMIC-III}
MIMIC-III (Medical Information Mart for Intensive Care) \cite{johnson2016mimic} is a large, publicly available critical care database comprising de-identified health records from over 58,000 hospital admissions at Beth Israel Deaconess Medical Center between 2001 and 2012. It includes data for 38,597 adult patients (aged $\geq$16 years) and 7,870 neonates.
The database provides detailed information on demographics, vital signs, laboratory results, medications, procedures, diagnostic codes (ICD-9), clinical notes, imaging reports, and outcomes such as mortality and length of stay. For each hospital admission, a median of 4,579 charted observations and 380 laboratory measurements are recorded.
Adult ICU patients have a median age of 65.8 years, with $55.9\%$ being male and an in-hospital mortality rate of $11.5\%$. The median ICU and hospital stays are 2.1 and 6.9 days, respectively. These characteristics make MIMIC-III a cornerstone dataset for tasks such as mortality prediction, readmission prediction, length-of-stay forecasting and  diagnosis Prediction.

\subsubsection{MIMIC-IV}
MIMIC-IV (Medical Information Mart for Intensive Care IV) \cite{johnson2023mimic} is a large-scale, publicly available electronic health record (EHR) dataset that captures de-identified health data from ICU and emergency department admissions at the Beth Israel Deaconess Medical Center between 2008 and 2019. Building upon its predecessors (MIMIC-II and MIMIC-III), MIMIC-IV expands both the temporal coverage and clinical detail, supporting more comprehensive and contemporary critical care research.
The dataset includes structured clinical data such as demographics, vital signs, laboratory results, diagnoses, procedures, medication administration (including precise eMAR records), and hospital-level metadata. It is organized into modular components to facilitate integration with other data modalities (e.g., imaging, waveforms) and external departmental sources.
MIMIC-IV complements other ICU datasets like eICU-CRD (multi-center), HiRID (high-frequency physiological signals), and AmsterdamUMCdb (European setting) by offering high-quality, single-center, time-resolved EHRs over an extended period. While it focuses primarily on structured clinical observations, MIMIC-IV also supports linking with free-text notes and future extensions to multi-modal data, addressing gaps in data diversity and clinical fidelity in ICU benchmarking resources.

\subsubsection{eICU}
eICU Collaborative Research Database (eICU-CRD) \cite{pollardEICUCollaborativeResearch2018} is a multi-center, high-resolution critical care dataset comprising de-identified electronic health records from 200,859 ICU stays across 208 U.S. hospitals between 2014 and 2015. It includes comprehensive time-stamped clinical data, such as vital signs, laboratory results, medications (including infusions), care plans, diagnoses, and interventions. Severity of illness scores (e.g., APACHE IV/IVa) and structured documentation from the eCareManager system enable robust risk stratification and outcome modeling.
The dataset spans 335 ICUs and provides hospital-level metadata (e.g., region, bed count, teaching status) and detailed patient demographics. Data tables cover diverse clinical domains, including intake/output volumes, microbiology results, and medication administration

\subsubsection{EHRShot}
EHRSHOT \cite{wornowEhrshotEhrBenchmark2023} is a large-scale benchmark dataset comprising de-identified, longitudinal electronic health records (EHRs) from 6,739 adult patients at Stanford Medicine. Unlike earlier datasets restricted to intensive care or emergency settings, EHRSHOT provides broader patient coverage across routine clinical care. It includes 41.6 million structured clinical observations across 921,499 encounters, encompassing demographics, diagnoses, procedures, laboratory results, and medications. The cohort excludes patients younger than 19 or older than 88 years, as well as those with fewer than 10 recorded clinical events.
The benchmark defines 15 clinically relevant classification tasks spanning operational outcomes (e.g., hospital length of stay), laboratory value prediction, future diagnosis assignment, and chest X-ray report classification.

\subsubsection{SymCat}
SymCat \cite{al-arsNLICESyntheticMedical2023} is a publicly available knowledge base that encodes probabilistic associations between symptoms and diseases based on large-scale aggregated EHR data. It includes 801 medical conditions and 474 symptoms, with condition–symptom relationships annotated by demographic priors over age, sex, and race.
For each condition, SymCat provides the likelihood of associated symptoms, along with demographic-specific probabilities:
(1) Age-based prevalence across eight groups (e.g., <1, 1–4, 5–14, …, >75 years);
(2) Sex-based distribution (male, female);
(3) Race-based distribution across White, Black, Hispanic, and Other populations.
Additionally, SymCat introduces structured symptom representations that capture richer clinical detail. For example, “pain” is modeled not just as present or absent, but with attributes such as intensity (mild, moderate, severe), location (chest, abdomen), and duration (acute, chronic). This enhancement supports the generation of more nuanced synthetic records and enables more realistic training and evaluation of differential diagnosis models.

\subsubsection{MedAlign}
MedAlign \cite{flemingMedalignCliniciangeneratedDataset2024} is a clinician-curated benchmark for evaluating the capability of language models to follow natural language instructions grounded in real-world electronic health records (EHRs). The dataset consists of 983 instructions, each linked to a de-identified longitudinal EHR, and spans a diverse set of clinically relevant tasks. Notably, MedAlign is released with a strict usage policy that prohibits its use for model pretraining or post-training.
Out of the 983 instructions, 303 include clinician-authored reference responses, supporting supervised evaluation. The remaining instructions are paired with 276 longitudinal EHRs, enabling model assessment in realistic, context-rich clinical settings.
Instructions are categorized into six functional groups:
(1) Retrieve $\&$ Summarize (e.g., summarizing annual physical exams)
(2) Care Planning (e.g., proposing treatment plans)
(3) Calculation $\&$ Scoring (e.g., computing stroke risk)
(4) Diagnosis Support
(5) Translation
(6) Administrative Prioritization.

\subsubsection{MedQA}
MedQA \cite{jinWhatDiseaseDoes2020} is a widely used single-turn QA benchmark derived from medical board exams in the United States (USMLE), Mainland China, and Taiwan. It evaluates LLMs’ ability to make accurate clinical decisions. Each example in the dataset includes a question either in one sentence asking for a certain piece of knowledge, or describing demographic details, symptoms, and relevant measurements, followed by multiple-choice options and the correct answer. In addition, supporting textual materials are provided, which can be leveraged to infer the correct answer. The questions span a wide range of clinical tasks, including diagnostic prediction, treatment recommendation, and prognostic prediction.

\subsubsection{Medbullets}
Medbullets \cite{chenBenchmarkingLargeLanguage2025} is a single-turn question answering (QA) dataset modeled after USMLE Step 2 and Step 3 exam formats. Each example consists of a clinical case description, a multiple-choice question with a ground-truth answer, and an annotated explanation, aiming to evaluate large language models’ (LLMs) capabilities in comprehensive clinical reasoning. The dataset contains 308 examples curated from the Medbullets online medical education platform. While Medbullets and MedQA share overlapping form, data source and medical domains, Medbullets is distinct in that it focuses exclusively on Step 2/3 questions, which are generally considered more complex and reasoning-intensive. Additionally, there is no data overlap between Medbullets and MedQA.

\subsubsection{HEAD-QA}
HEAD-QA \cite{vilaresHEADQAHealthcareDataset2019} is a single-turn QA dataset designed to evaluate the reasoning capabilities of LLMs. It consists of graduate-level, multiple-choice medical questions across several disciplines, including nursing, biology, chemistry, psychology, and pharmacology. The dataset includes 6.7k supervised and 6.7k unsupervised examples. A subset of the data also includes images as an additional modality.

\subsubsection{Open-XDDx}
Open-XDDx \cite{zhouExplainableDifferentialDiagnosis2025} is a well-annotated dataset developed for explainable differential diagnosis (DDx), comprising 570 clinical notes derived from publicly available medical education exercises. These notes span nine clinical specialties, including cardiovascular, digestive, respiratory, endocrine, nervous, reproductive, circulatory, skin, and orthopedic disorders. Each note includes structured patient symptoms, corresponding differential diagnoses, and expert-derived explanations, making the dataset well-suited for studying interpretable and multi-step diagnostic reasoning.
On average, each note contains 4.6 diagnoses and 14.5 explanations, with 3.1 explanations per diagnosis. This level of granularity enables evaluation of both diagnostic accuracy and the traceability of model-generated reasoning.

\subsubsection{DDXPlus}
DDXPlus \cite{fansitchangoDdxplusNewDataset2022} is a large-scale, synthetic benchmark dataset developed to support differential diagnosis (DDx) modeling and explainability in clinical NLP. It comprises approximately 1.3 million simulated patient cases, each associated with one ground-truth pathology and a list of ranked differential diagnoses. The dataset is generated using a proprietary rule-based assistant diagnostic (AD) system built from a curated knowledge base of over 20,000 medical articles, reviewed by clinicians.
Each patient case includes demographic attributes (age, sex, and geographic information), symptoms, antecedents, and fine-grained labels such as symptom types (binary, categorical, multi-choice), severity levels, and hierarchical symptom relationships. Notably, symptoms and antecedents are explicitly distinguished and modeled with conditional dependencies. A key innovation of DDXPlus is the inclusion of both true diagnoses and rich differential lists, enabling models to learn probabilistic ranking and evidential reasoning.
DDXPlus supports downstream tasks including differential diagnosis ranking, evidence collection planning, interpretable reasoning, and active diagnostic dialogue modeling.

\subsubsection{MEDEC}
MEDEC \cite{abacha2025medecbenchmarkmedicalerror} is a free-text dataset designed for medical error detection, consisting of 3,848 clinical texts. Among these, 488 were sourced from three U.S. hospital systems that had not been previously accessed by any large language models (LLMs). The dataset was constructed by manually injecting medical errors into two sources: question contexts from MedQA \cite{jinWhatDiseaseDoes2020} and real-world clinical notes. The injected errors span five categories: diagnosis, causal organism, management, treatment, and pharmacotherapy.

\subsubsection{MedCalc-Bench}
MedCalc-Bench \cite{khandekar2024medcalc} is a single-turn question answering(QA) dataset aimed at evaluating the quantitative clinical reasoning abilities of large language models. It consists of over one thousand manually reviewed instances covering 55 distinct types of medical calculation tasks. Each instance includes a patient note as context, a single-turn question requiring a numerical medical calculation, a ground-truth answer, and corresponding explanation.

\subsubsection{PubMedQA}
PubMedQA \cite{jin2019pubmedqa} is a single-turn QA dataset designed for medical research assistance, constructed from PubMed abstracts. Each instance in the dataset consists of: (1) a medical research question derived from a biomedical research paper; (2) the abstract of that paper as context; (3) the conclusion sentence of the abstract as the Long Answer; and (4) a categorical label indicating whether the Long Answer address the question with yes/no/maybe. The task for language models is to predict the correct judgment label based on the context. The dataset includes 1k expert-annotated, 61.2k unlabeled, and 211.3k artificially generated QA instances. In addition to medical research assistance, PubMedQA has also been used in diagnosis prediction tasks \cite{kimMdagentsAdaptiveCollaboration2024}.

\subsubsection{MedHallu}
MedHallu \cite{panditMedHalluComprehensiveBenchmark2025} is a benchmark dataset specifically designed for medical hallucination detection. It consists of 10,000 QA pairs derived from PubMedQA \cite{jin2019pubmedqa}, each paired with a hallucinated answer generated through a predefined pipeline. The dataset is categorized into three difficulty levels based on the challenge of detecting hallucinations. The task requires LLMs to classify whether an answer is hallucinated. Hallucinations in MedHallu fall into four categories: (1) misinterpretation of the question, (2) incomplete information, (3) misattribution of biological mechanisms or pathways, and (4) fabrication of methodology or evidence. Although EHR features are not directly meaningful in this task, we retain them in Table \ref{tab:dataset_comparison} for consistency, as the context originates from the same source as PubMedQA.

\subsubsection{MTSamples}
MTSamples \cite{mtsamples} is an online collection of user-uploaded or transcriptionist-generated medical reports, organized in free-text form. The dataset spans over 40 medical specialties, such as Allergy, Autopsy, and Bariatrics, and currently contains more than 5,000 records. Notably, some of the samples include partial differential diagnosis information, often indicated by keywords such as “preliminary diagnosis” or “differential diagnosis” within the report text.

\subsubsection{DischargeMe}
DischargeMe \cite{xuOverviewFirstShared2024} is a free-text clinical text generation dataset derived from the MIMIC-IV-Note module \cite{johnson2023mimicivnote}. It requires large language models (LLMs) to generate discharge instructions and brief hospital course summaries based on input text such as radiology reports. The dataset comprises 109,168 hospital admissions originating from the Emergency Department. Each visit includes chief complaints and diagnosis codes, at least one radiology report, and a discharge summary.

\subsubsection{ACI-Bench}
ACI-Bench \cite{yim2023aci} is a free-text dataset for LLM-based clinical note generation, featuring paired patient–doctor conversations and structured clinical notes. It comprises 207 synthetic patient encounters, created based on three real-world clinical documentation paradigms: virtual assistant, virtual scribe, and ambient clinical intelligence.

\subsubsection{MedicationQA}
MedicationQA \cite{abachaBridgingGapConsumers2019} is a single-turn QA dataset designed to address medication-related queries from consumers. It contains over 674 QA pairs, each comprising a real-world consumer question about medication collected on MedlinePlus \cite{medlineplus}, along with manually annotated information including the question focus (always the drug name), question type (e.g., Drug Information, Dosage, Side Effects), and manually retrieved answer sources.


\input{Tables/Experiments/event_num}

\input{Sections/Experimental_design/data_property_comparison}

\input{Sections/Experimental_design/evaluation_metric}

%% file: Tables/Experiments/event_num.tex

\begin{table*}[ht]
\centering
\small
\begin{tabular}{c|c|p{3.75 cm}|p{3.75 cm}|p{4.2 cm}}
\hline
\textbf{Dataset Name} & \textbf{Size} & \textbf{Diseases} & \textbf{Data Source} & \textbf{Tasks}\\
\hline
MIMIC-III   & 53K             & ICU admissions                                  & Beth Israel Deaconess Medical Center (BIDMC) & General-purpose tabular data \newline (e.g. Mortality Prediction, Treatment Effect Estimation) \\
MIMIC-IV    & 65K             & ICU and ED admissions                           & Beth Israel Deaconess Medical Center (BIDMC) & General-purpose tabular data (e.g. Mortality Prediction, Risk Prediction) \\
eICU-CRD     & 139K            & ICU-focused (multi-site)                        & 208 hospitals located throughout the US & General-purpose tabular data (e.g. Mortality Prediction, Risk Prediction) \\
EHRSHOT       & 7K              &  Highly heterogeneous clinical diagnoses beyond ICU/ED admissions & Stanford’s STARR-OMOP database & General-purpose tabular data, 15 predefined downstream tasks \newline(e.g. Readmission Prediction, Diagnosis Prediction)\\
SymCat        & --                 & 801 diseases across primary care                & Synthetic & Diagnosis Prediction\\
\midrule
MedAlign      & 276                & Highly heterogeneous clinical diagnoses beyond ICU/ED admissions & Stanford’s STARR-OMOP database & General-purpose tabular + QA data, 5 predefined downstream tasks \newline(e.g. Clinical Note Generation)\\
MedQA       & 61K              &  Various diseases featured in the professional medical board exams, e.g. US Medical Licensing Examination (USMLE)  & Professional medical board exams from the US (USMLE), Mainland China, and Taiwan. & Comprehensive clinical decision support \newline(e.g. Diagnosis Prediction, Treatment Recommendation)\\
Medbullets       & 308          & Various diseases featured in the professional medical board exams & Online platform Medbullets, Step 2/3 USMLE style questions & Comprehensive clinical decision support\\
HeadQA       & 6.7K    & Diverse diseases across nursing, pharmacology, medicine, and psychology & Questions from Spanish healthcare specialization exams & Comprehensive clinical  decision support \\
Open-XDDx     & 570           & 9 disease categories, including cardiovascular, digestive, respiratory, endocrine, and others & Medical books and MedQA USMLE dataset & Differential Diagnosis Prediction\\
DDXPlus       & 1.3M               & Various simulated diagnoses              & Synthetic & Differential Diagnosis Prediction\\
MEDEC       & 4K              &  Task spans 5 clinical processes, including diagnoses from diverse areas & Annotated MedQA \& Three University of Washington (UW) hospital systems & Medical Error Detection\\
MedCalc-Bench       & 1K        & Cardiovascular, cerebrovascular, renal, hepatic, metabolic, respiratory, and thrombotic diseases  & MDCalc & Quantitative Clinical Calculations\\
PubMedQA       & 273K              & Heterogeneous diseases mentioned in PubMed articles & Abstracts from PubMed articles, with QA annotated manually or automatically & Concept Extraction, \newline Diagnosis Prediction\\
MedHallu       & 10k              &  Heterogeneous diseases mentioned in PubMed articles & Generated based on PubMedQA dataset & Concept Extraction, \newline Medical Hallucination Detection\\
MTSamples       & 5K              &  Diagnoses spanning 40 medical specialties & User-contributed samples from MTSamples.com & Concept Extraction,\newline Treatment Recommendation\\
DischargeMe       & --            &  ED admissions, radiology & MIMIC-IV Note Module & Clinical Note Generation\\
ACI-Bench       & 207             &  Various simulated diagnoses & Synthetic, human annotated & Clinical Note Generation\\
MedicationQA       & 674           &  Diseases potentially associated with drug administration (e.g. drug-induced allergies) & Anonymized consumer questions submitted to MedlinePlus & Clinical Value (Medication) \newline Interpretation\\
\hline
\end{tabular}
\caption{Statistical summary of clinical datasets, including patient counts, disease coverage, data sources, and downstream tasks. Note that dataset size refers to the number of patients for structured/tabular EHR data, and to the number of records or instances for free-text and single-turn QA datasets.}
\label{tab:dataset_stats}
\end{table*}

%% file: Sections/Experimental_design/data_property_comparison.tex

\begin{table*}[ht]

\centering
\small
\begin{tabular}{l|l|l|*{5}{c}}
\hline
\multirow{2}{*}{\textbf{Dataset Name}} & \multirow{2}{*}{\textbf{Format}} & \multirow{2}{*}{\textbf{Extra Modalities}} & \multicolumn{5}{c}{\textbf{EHR Features}} \\
\cline{4-8}
& & & \textbf{Demographics} & \textbf{Symptoms} & \textbf{Timestamps} & \textbf{Lab Info} & \textbf{DDx} \\
\hline
MIMIC-III     & Tabular         & Text             & \cmark & \xmark & \cmark & \cmark & \xmark \\
MIMIC-IV      & Tabular         & Text             & \cmark & \xmark & \cmark & \cmark & \xmark \\
eICU-CRD      & Tabular         & --               & \cmark & \xmark & \cmark & \cmark & \xmark \\
SymCat        & Tabular         & --           & \cmark & 474    & \xmark & \xmark & \cmark \\
EHRSHOT       & Tabular         & --           & \cmark & \cmark & \cmark & \cmark & \xmark \\
\midrule
MedAlign      & Single-turn QA  & Structured EHR   & \cmark & \cmark & \cmark & \cmark & \xmark \\
MedQA       & Single-turn QA         & --           & \cmark & \cmark & \omark & \omark & \xmark \\
Medbullets       & Single-turn QA         & --           & \cmark & \cmark & \omark & \cmark & \xmark \\
HeadQA       & Single-turn QA         & Image           & \cmark & \cmark & \xmark & \omark & \xmark \\
Open-XDDx     & Single-turn QA  & --               & \cmark & \cmark & \omark & \omark & \cmark \\
DDXPlus       & Single-turn QA  & --               & \cmark & 110    & \omark & \xmark & \cmark \\
MEDEC       & Free-text        & --           & \cmark & \cmark & \omark & \omark & \xmark \\
MedCalc-Bench      & Single-turn QA         & --           &\cmark  & \xmark & \cmark & \cmark & \xmark \\
PubMedQA       & Single-turn QA & --           & \omark & \cmark & \xmark & \omark & \xmark \\
MedHallu     & Single-turn QA         & --           & \omark & \cmark & \xmark & \omark & \xmark \\
MTSamples       & Free-text         & --          & \cmark & \cmark & \xmark & \xmark & \omark \\
DischargeMe       & Free-text         & Structured EHR   & \cmark & \cmark & \cmark & \cmark & \xmark \\
ACI-Bench   & Free-text         & --           & \cmark & \cmark & \xmark & \xmark & \xmark \\
MedicationQA  & Single-turn QA         & --    & \xmark & \omark & \xmark & \xmark & \xmark \\
\hline
\end{tabular}
\caption{Comparison of Clinical Datasets by Format, Modalities, and EHR Features. Note that \textbf{text} is the inherent modality of Free-text or QA datasets, while \textbf{structured EHR} is the inherent modality of tabular datasets. As such, these modalities are not explicitly annotated in the table. \omark\ indicates that the dataset partially or unstructured contains this type of information.}
\label{tab:dataset_comparison}
\end{table*}

%% file: Sections/Experimental_design/evaluation_metric.tex
\subsection{Evaluation Metrics in EHR Prediction}

Predictive modeling using Electronic Health Records (EHR) involves diverse tasks such as classification, regression, survival analysis, time series forecasting, and generative modeling. Evaluating models rigorously is essential for clinical translation and safety. AA summary of the metrics and their formulas is shown in Table \ref{tab:metric}.

\begin{table}[htbp]
\centering
\resizebox{0.85\textwidth}{!}{%
\renewcommand{\arraystretch}{1.4}
\begin{tabular}{@{}lllp{6cm}@{}}
\toprule
\textbf{Task} & \textbf{Aspect} & \textbf{Metric} & \textbf{Formula} \\ 
\midrule

\multirow{4}{*}{Classification} 
  & - & Accuracy & $\displaystyle \frac{TP + TN}{TP + TN + FP + FN}$ \\
  & - & Precision & $\displaystyle \frac{TP}{TP + FP}$ \\
  & - & Recall & $\displaystyle \frac{TP}{TP + FN}$ \\
  & - & F1-score & $\displaystyle 2 \times \frac{\text{Precision} \times \text{Recall}}{\text{Precision} + \text{Recall}}$ \\
\midrule

\multirow{4}{*}{Regression} 
  & - & MAE & $\displaystyle \frac{1}{n} \sum_{i=1}^n |y_i - \hat{y}_i|$ \\
  & - & MSE & $\displaystyle \frac{1}{n} \sum_{i=1}^n (y_i - \hat{y}_i)^2$ \\
  & - & RMSE & $\displaystyle \sqrt{\text{MSE}}$ \\
  & - & $R^2$ & $\displaystyle 1 - \frac{\sum (y_i - \hat{y}_i)^2}{\sum (y_i - \bar{y})^2}$ \\
\midrule

\multirow{2}{*}{Time-to-Event Forecasting} & - & C-index & $\displaystyle \frac{\text{Concordant Pairs}}{\text{Comparable Pairs}}$ \\
& - & Integrated Brier Score & $\text{IBS} = \int_0^\tau BS(t) \, dt$ \\

\midrule

\multirow{4}{*}{Time Series Forecasting Metrics} 
  & - & Step-wise MAE & $\displaystyle \frac{1}{n} \sum_{i=1}^n |y_i - \hat{y}_i|$ \\
  & - & Step-wise RMSE & $\displaystyle \sqrt{ \frac{1}{n} \sum_{i=1}^n (y_i - \hat{y}_i)^2 }$ \\
  & - & Dynamic Time Warping (DTW) & $\displaystyle \min_{p \in P} \sum_{(i,j) \in p} d(x_i, y_j)$ \\
  & - & Prediction Interval Coverage & $\displaystyle \frac{1}{n} \sum_{i=1}^n \mathbf{1}[y_i \in \hat{I}_i]$ \\
\midrule

\multirow{9}{*}{Ranking Metrics} 
  & coverage & Coverage Error & $\displaystyle \frac{1}{n} \sum_{i=1}^n \max_{j \in Y_i} \text{rank}_i(j) - 1$ \\
  & coverage & Ranking Loss & $\displaystyle \frac{1}{n} \sum_{i=1}^n \frac{1}{|Y_i||\overline{Y}_i|} \sum_{(j,k)} \mathbf{1}[\text{rank}_i(j) > \text{rank}_i(k)]$ \\
  & coverage & One-Error & $\displaystyle \frac{1}{n} \sum_{i=1}^n \mathbf{1}[\text{top}_i \notin Y_i]$ \\
  & coverage & Recall@k & $\displaystyle \frac{1}{n} \sum_{i=1}^n \frac{|Y_i \cap \hat{Y}_{i,k}|}{|Y_i|}$ \\
  & early prediction & Precision@k & $\displaystyle \frac{1}{n} \sum_{i=1}^n \frac{|Y_i \cap \hat{Y}_{i,k}|}{k}$ \\
  & early prediction & Hit Rate@k & $\displaystyle \frac{1}{n} \sum_{i=1}^n \mathbf{1}[Y_i \cap \hat{Y}_{i,k} \neq \emptyset]$ \\
  & early prediction & Mean Reciprocal Rank & $\displaystyle \frac{1}{n} \sum_{i=1}^n \frac{1}{\text{rank}_i}$ \\
  & early prediction & Average Precision & $\displaystyle \sum_k (R_k - R_{k-1}) P_k$ \\
  & early prediction & NDCG@k & $\displaystyle \frac{1}{n} \sum_{i=1}^n \frac{DCG_i@k}{IDCG_i@k}$ \\
\midrule

\multirow{2}{*}{Generative Modeling Metrics} 
  & - & Dynamic Time Warping (DTW) & $\displaystyle \min_{P} \sum_{(i,j) \in P} d(x_i, y_j)$ \\
  & - & Maximum Mean Discrepancy (MMD) & $\displaystyle \left\| \frac{1}{n} \sum_{i=1}^n \phi(x_i) - \frac{1}{m} \sum_{j=1}^m \phi(y_j) \right\|^2$ \\

\bottomrule
\end{tabular}
}
\caption{Evaluation Metrics on different Downstream Tasks.}
\label{tab:metric}
\end{table}

\subsubsection{Classification Metrics}

Classification tasks in EHR prediction include binary outcomes (e.g., readmission prediction, multiclass problems (e.g., disease subtype classification, clinical test recommendation) and multilabel problems (e.g., comorbidities prediction). Standard metrics include: Accuracy measures the fraction of correct predictions among all samples. 
Precision measures the proportion of true positives $TP$ among positive predictions $TP + FP$, while recall measures the proportion of true positives $TP$ among actual positives $TP + FN$. The F1-score balances both.
Area Under ROC Curve (AUROC) summarizes the trade-off between true positive rate (TPR) and false positive rate (FPR) across thresholds. Area Under Precision-Recall Curve (AUPRC) better reflects model performance when the positive class is rare.
In EHR prediction, multilabel classification is used when patients can have multiple simultaneous diagnoses (e.g., comorbidities). Common metrics include Micro, Macro, and Weighted F1-scores.

\subsubsection{Regression Metrics}
For continuous outcomes such as length of stay or lab values, common metrics include:
Mean Absolute Error (MAE) measures average absolute deviation. Mean Squared Error (MSE) and Root Mean Squared Error (RMSE) penalize larger errors more heavily.
R-squared ($R^2$) explains variance.

\subsubsection{Time-to-Event Prediction (Survival Analysis)}

Time-to-event prediction is central to predicting time until an event (e.g., death, disease progression).
Concordance Index (C-index) evaluates how well predicted risk ranks survival times. A value of 0.5 indicates random predictions; 1.0 indicates perfect concordance.
Integrated Brier Score combines discrimination and calibration over time via calculating the integration of the time-dependent Brier score $BS(t)$.
Time-dependent AUROC calculates AUROC at specific time horizons, capturing time-varying discrimination.

\subsubsection{Time Series Forecasting Metrics}

Predicting future vital signs or lab trends requires sequential evaluation:
Step-wise Error measures MAE and RMSE at each forecast step. Dynamic Time Warping (DTW) measures alignment of entire sequences following a warping path $P$. Forecasting Horizon Error aggregates error over multiple time steps, reflecting error propagation. Prediction Interval Coverage quantifies uncertainty by evaluating whether true values fall within predicted intervals.

\subsubsection{Ranking Metrics}
Ranking-based metrics evaluate the quality of predicted label or item rankings in EHR tasks, such as differential diagnosis, medication recommendation, or risk stratification. These metrics are typically divided into two categories: coverage and early prediction. Coverage metrics (e.g., Coverage Error, Ranking Loss, One-Error, Recall@k) measure how well the model ensures that all relevant clinical labels or items are eventually included in the ranked list, emphasizing the ability to comprehensively capture true positives. Early Prediction metrics (e.g., Precision@k, Hit Rate@k, Mean Reciprocal Rank, Average Precision, Normalized Discounted Cumulative Gain (NDCG), which compares the predicted ranking to the ideal ranking by normalizing the Discounted Cumulative Gain ($DCG$) with the Ideal DCG ($IDCG$)) focus on ranking the most relevant items as early as possible, reflecting the clinical need to surface the most critical or likely diagnoses and recommendations at the top of the list. This distinction ensures that models are evaluated both on their ability to retrieve all necessary information and on their prioritization of the most actionable results.

\subsubsection{Generative Modeling Metrics}

Evaluating generative modeling in EHR research requires assessing how well synthetic data replicate real data distributions, maintain clinical validity, and protect patient confidentiality. Common metrics include n-gram overlap measures such as BLEU, ROUGE, and METEOR, which evaluate text generation quality in clinical notes by comparing precision and recall of n-grams, though they may miss deeper clinical meaning. For time series generation (e.g., vital signs), Dynamic Time Warping (DTW) quantifies sequence alignment, while Maximum Mean Discrepancy (MMD) assesses distributional similarity in a reproducing kernel Hilbert space. Classifier Two-sample Tests train discriminators to distinguish real from synthetic samples, with poor discrimination indicating better fidelity. Expert review remains essential for evaluating clinical plausibility, ensuring that generated records are realistic and medically consistent. Additionally, for open-ended text generation tasks, LLM-jury ensemble methods use multiple language models to score outputs on Likert scales for medical accuracy, completeness, and clarity, complemented by automated metrics such as ROUGE and BERTScore to capture lexical and semantic overlap. Together, these metrics provide a comprehensive framework for evaluating both statistical similarity and clinical utility of synthetic EHR data.












%% file: Sections/future_works.tex
\section{Emerging Trends and Open Problems}
\label{sec::challenge_future}
\subsection{Emerging Trend}
\textbf{Foundation Models for EHRs}. Clinical LLMs have achieved strong performance on medical question answering benchmarks, typically trained on text from clinical literature, textbooks, or social media. However, these sources differ substantially from real-world EHRs, which are structured, heterogeneous, and domain-specific.
Modeling such data requires both numerical reasoning and contextual understanding of clinical semantics. LLMs trained solely on unstructured text using next-token prediction often struggle to interpret structured inputs and lack grounding in domain-specific medical knowledge.
Transforming structured EHRs into LLM-compatible formats is emerging as a key direction for enabling scalable and clinically meaningful reasoning.
Recently, DiaLMs \cite{ren2025diallms} introduced the first framework to translate heterogeneous EHR data into clinically meaningful text using a manually curated Clinical Test Reference (CTR). The CTR facilitates the translation of standardized codes (e.g., ICD-9/10, LOINC) into interpretable clinical language and enables context-aware interpretation of numerical test results, conditioned on patient-specific attributes such as age and gender.

\textbf{Foundation Models for Multi-modal Clinical Data}. 
Since 2022, most existing efforts have focused on developing general-purpose multimodal vision-language models (VLMs) to answer broad clinical questions across diverse data modalities. However, clinical imaging varies significantly in resolution, modality, and diagnostic intent, and domain-specific knowledge is often not easily transferable across tasks. These limitations pose challenges for generalist models in high-stakes clinical settings.
To address this, recent research has begun to explore disease-specific multimodal models that integrate imaging, structured EHRs, and patient-specific clinical context. These systems aim to provide personalized diagnostic reasoning and treatment planning, tailored to particular clinical domains such as oncology, ophthalmology, or cardiology. Such specialization improves interpretability, aligns better with real-world workflows, and enhances performance on narrowly scoped, clinically meaningful tasks.

\textbf{Clinical Agent}. Agent-based systems provide a structured framework for coordinating multiple clinical tasks within a unified workflow. Recent studies have introduced agent architectures that simulate clinical decision-making processes, such as hypothesis generation and multi-expert consultation. These agents support tasks including differential diagnosis, test recommendation, and personalized care planning.
Agent-based systems provide a structured framework for coordinating multiple clinical tasks within a unified workflow. Recent studies have introduced agent architectures that simulate clinical decision-making processes, including hypothesis generation, information retrieval, and multi-expert consultation. These agents support tasks such as differential diagnosis, test recommendation, and personalized care planning.
Although agent systems have shown strong capabilities in general-purpose applications, such as navigating web environments, querying external databases, and executing multi-step reasoning, their use in clinical contexts remains limited. Building clinically grounded agents that can interact with structured EHRs, incorporate domain-specific knowledge, and support real-time collaboration with healthcare providers presents an important future direction for clinical AI research.

\textbf{Embodied Clinical AI}.
Recent advances in artificial intelligence have significantly improved capabilities in perception, forming the basis for embodied clinical AI systems that operate in physical healthcare environments. Applications such as surgical robotics, smart ICU monitoring, and bedside assistance demonstrate the potential of integrating perception with decision-making and control. This integration enables real-time, interactive, and context-aware clinical support, expanding the role of AI from passive analysis to active participation in patient care.

\subsection{Open problems}

\subsubsection{Benchmark and validation}

\textit{Data Collection}.
High-quality, large-scale EHR datasets are critical for training and benchmarking clinical models. However, real-world clinical data are fragmented across institutions, subject to strict privacy regulations, and highly variable in structure and annotation. Most publicly available datasets (e.g., MIMIC, eICU) are limited in scope, demographic diversity, and task coverage, which hinders generalizability and reproducibility. There is a pressing need for more representative, longitudinal, and task-aligned datasets that reflect diverse patient populations and clinical settings.

\textit{Data Synthesis}.
Synthetic data generation has emerged as a potential solution to data access and privacy barriers. However, generating clinically valid and temporally coherent EHR data remains a challenge. Existing generative models often struggle with maintaining semantic consistency, preserving rare but critical clinical patterns, and replicating the hierarchical and multi-modal structure of real-world records. Without rigorous validation, synthetic data may misrepresent clinical realities and lead to misleading model performance estimates.

\textit{Evaluation}.
Evaluation practices in clinical AI remain fragmented and insufficiently aligned with real-world clinical needs. Most studies rely on general-purpose metrics such as accuracy, AUC, or F1 score, which fail to capture clinical relevance, safety, or decision impact. Furthermore, current benchmarks often focus on single-institution datasets and narrow tasks, limiting the assessment of model generalizability across patient populations, care settings, and time periods. Critical dimensions such as fairness, robustness, uncertainty calibration, and interpretability are rarely incorporated into standard evaluation pipelines. To enable trustworthy deployment, future benchmarks should adopt multi-dimensional evaluation frameworks that reflect clinical utility, ensure reproducibility, and support transparent model comparisons across modalities and tasks.

\subsubsection{Modeling Structured and Temporal EHRs.}
While numerous studies have proposed architectural and representation strategies to enhance empirical performance on electronic health record (EHR) modeling tasks, effectively capturing the heterogeneous, hierarchical, and temporally irregular nature of EHR data remains an open challenge. Most existing approaches rely on transformer-based architectures, which may not fully leverage these structural characteristics. EHRs can be naturally represented as sequences of structured tables, exhibiting hierarchies across visits, events, and clinical codes. Tree-based models, in particular, are well suited for capturing such dependencies and have shown strong performance in patient representation learning and longitudinal prediction. Future research should focus on developing specialized architectures that explicitly incorporate EHR-specific heterogeneity and temporal dynamics to enhance model interpretability and support accurate long-term clinical forecasting.

\subsubsection{Explainability and Alignment with Clinical Knowledge}. 
Despite recent improvements in predictive performance, the adoption of AI systems in healthcare requires more than accuracy alone. Clinical settings demand transparency, trustworthiness, and alignment with domain-specific standards and workflows. Several key challenges remain:

(1) \textit{Alignment with Clinical Workflow}: Unlike general AI applications (e.g., recommendation or image classification), clinical decision-making is inherently interactive and iterative. Most existing models focus narrowly on diagnosis assistance using simplified "symptom-to-diagnosis" data. However, real-world diagnostics involve repeated cycles of evidence gathering, test ordering, and interpretation, all recorded in EHRs. Simulating this dynamic decision process within LLMs remains an open challenge. Moreover, AI systems should extend beyond diagnosis to support broader clinical workflows, including administrative triage, resource planning, and operational management. This calls for outputs that are interpretable, timely, and actionable within real-world healthcare systems.

(2) \textit{Alignment with Clinical Language and Guidelines}: To ensure safety and avoid legal or ethical concerns, model outputs must conform to established clinical terminology, official guidelines, and institutional standards. Alignment with these domain constraints is essential for trust, regulatory compliance, and safe clinical deployment.

(3) \textit{Interpretability and Clinical Reasoning}: Accurate predictions alone are insufficient. Clinical AI systems must offer transparent reasoning, provide uncertainty estimates, and generate warnings when needed. Supporting comprehensive clinical reasoning is crucial for fostering clinician trust, improving decision support, and enabling safe use in high-stakes environments.